\pdfoutput=1

\documentclass[11pt]{article}

\usepackage[table]{xcolor}  
\usepackage[final]{acl}

\usepackage{times}
\usepackage{latexsym}

\usepackage[T1]{fontenc}

\usepackage[utf8]{inputenc}

\usepackage{microtype}

\usepackage{inconsolata}

\usepackage{graphicx}
\newcommand{\letsc}{\texttt{LETS-C}}

\usepackage{newfloat}
\usepackage{listings}
\usepackage[utf8]{inputenc} 
\usepackage[T1]{fontenc}    
\usepackage{url}            
\usepackage{booktabs}       
\usepackage{amsfonts}       
\usepackage{nicefrac}       
\usepackage{microtype}      
\usepackage{enumitem}
\usepackage{graphicx}
\usepackage{adjustbox}
\usepackage{multirow}
\usepackage{amsmath}
\usepackage{makeidx} 
\usepackage{appendix} 
\usepackage{titletoc}
\usepackage{floatflt}

\usepackage{algorithm}
\usepackage{algorithmic}

\usepackage{lipsum}
\usepackage{graphicx}

\colorlet{pale1}{blue!10}
\colorlet{pale2}{green!10}
\colorlet{pale3}{red!10}
\colorlet{pale4}{orange!10}
\colorlet{pale5}{cyan!10}
\colorlet{pale6}{magenta!10}
\colorlet{pale7}{gray!10}
\colorlet{pale8}{teal!10}
\colorlet{pale9}{purple!10}
\newcolumntype{P}[1]{>{\centering\arraybackslash}p{#1}}

\title{LETS-C: Leveraging Text Embedding for Time Series Classification}

\author{%
Rachneet Kaur \quad Zhen Zeng \quad
Tucker Balch \quad Manuela Veloso \\
J.P. Morgan AI Research\\
\texttt{\{rachneet.kaur, zhen.zeng, tucker.balch, manuela.veloso\}@jpmorgan.com}
}

\begin{document}
\maketitle
\begin{abstract}
Recent advancements in language modeling have shown promising results when applied to time series data. In particular, fine-tuning pre-trained large language models (LLMs) for time series classification tasks has achieved state-of-the-art (SOTA) performance on standard benchmarks. However, these LLM-based models have a significant drawback due to the large model size, with the number of trainable parameters in the millions.
In this paper, we propose an alternative approach to leveraging the success of language modeling in the time series domain. Instead of fine-tuning LLMs, we utilize a text embedding model to embed time series and then pair the embeddings with a simple classification head composed of convolutional neural networks (CNN) and multilayer perceptron (MLP).
We conducted extensive experiments on a well-established time series classification benchmark. We demonstrated LETS-C not only outperforms the current SOTA in classification accuracy but also offers a lightweight solution, using only 14.5\% of the trainable parameters on average compared to the SOTA model. 
Our findings suggest that leveraging text embedding models to encode time series data, combined with a simple yet effective classification head, offers a promising direction for achieving high-performance time series classification while maintaining a lightweight model architecture.
\end{abstract}

\section{Introduction}\label{sec:intro}
Time series classification~\cite{bagnall2017great, abanda2019review, ismail2019deep} has gained attention due to its wide-ranging applications in various domains, such as finance~\cite{passalis2017time}, healthcare~\cite{lipton2016directly}, and activity recognition~\cite{yang2015deep}. The increasing availability of time series data has driven the need for efficient and accurate classification methods. Advances in natural language processing (NLP) and large language models (LLMs)~\cite{achiam2023gpt} have demonstrated strong promises in modeling sequential data~\cite{achiam2023gpt}. Inspired by this, researchers have explored applying LLMs to time series via prompting~\cite{gruver2024large, liu-etal-2024-lstprompt, merrill-etal-2024-language, chu-etal-2024-timebench} or fine-tuning pre-trained LLMs~\cite{jin2023time, zhou2024one}, achieving state-of-the-art (SOTA) performance on well-established benchmarks for tasks including classification and forecasting.

However, LLMs for time series classification face a major drawback—their large size, often with billions of parameters, making them computationally expensive and impractical in resource-limited settings~\cite{bommasani2021opportunities}. Fine-tuning partially frozen pre-trained LLMs still requires millions of trainable parameters~\cite{zhou2024one}. To mitigate this, we propose an alternative approach to leverage the success of language modeling in the time series domain. 
In particular, we propose a novel approach, \letsc{} (\textbf{L}everaging Text \textbf{E}mbeddings for \textbf{T}ime \textbf{S}eries \textbf{C}lassification), which utilizes off-the-shelf text embedding models instead of fine-tuning LLMs for time series classification.
To the best of our knowledge, this work is the \underline{first} to explore the potential of text embeddings in time series analysis, specifically classification, and demonstrate SOTA performance.

\letsc{} combines text embeddings with a simple yet effective classification head composed of convolutional neural networks (CNNs) and a multilayer perceptron (MLP). By projecting time series data using text embedding models, we capture the intricate patterns and dependencies present in the temporal data. The embeddings and time series are then fed into the classification head, which learns to discriminate between different classes. Through extensive experiments on a well-established benchmark containing time series datasets from various domains, we demonstrated that \letsc{} outperforms 27 baselines including the previous SOTA method. Moreover, \letsc{} is significantly more efficient, using much fewer trainable parameters than the previous SOTA. 
Our key contributions are:
\begin{itemize}[leftmargin=*, noitemsep]
    \item \textbf{\textit{Text Embeddings for Time Series:}} We introduce \letsc{}, the \underline{first} work to leverage text embeddings for time series analysis, specifically for classification tasks.
    \item \textbf{\textit{State-of-the-Art Performance:}} \letsc{} achieves SOTA performance in classification accuracy on a well-established benchmark containing time series datasets across different domains, surpassing 27 baseline models.
    \item \textbf{\textit{Computationally Lightweight:}} \letsc{} is significantly more efficient, achieving higher accuracy while using much fewer trainable parameters (14.5\%) than the existing SOTA method.
\end{itemize}
In addition, we conducted comprehensive analyses to showcase the effectiveness of \letsc{}:
\begin{itemize}[leftmargin=*, noitemsep]
    \item \textbf{\textit{Built-in Discriminative Power of Time Series Embeddings:}} We demonstrate the advantage of text embeddings for time series, showing that embeddings of time series from the same class are more similar than those from different classes, explaining the boost in classification accuracy.
    
    \item \textbf{\textit{Generalization Across Various Text Embedding Models:}} \letsc{} with different text embedding models consistently outperforms previous SOTA with much fewer trainable parameters, further validating the effectiveness of our approach.
    
    \item \textbf{\textit{Optimizing Model Size with Minimal Accuracy Loss:}} \letsc{} retains a high percentage of accuracy even as the  classification model size shrinks considerably, enhancing computational efficiency with minimal impact on performance.
\end{itemize}
\section{Related Work}\label{sec:related_works}
We review the related works in three key areas: time series classification, the application of language models to time series data, and text embeddings. See Appendix \ref{appendix:related_works} for an extended review.

\paragraph{Time Series Classification}
Time series classification has been an active research area for decades. Early methods focused on distance-based approaches~\cite{abanda2019review}, such as Dynamic Time Warping~\cite{berndt1994using} and distance kernels with Support Vector Machines ~\cite{kampouraki2008heartbeat}. Others used feature extraction with classifiers like eXtreme Gradient Boosting (XGBoost)~\cite{chen2016xgboost} and LightGBM~\cite{ke2017lightgbm}. 
Deep learning-based approaches, including CNNs~\cite{wu2022timesnet, zhao2017convolutional}, MLPs~\cite{zhang2022less}, and Recurrent Neural Networks (RNNs) like Long Short-Term Memory (LSTM)~\cite{lai2018modeling}, later gained popularity for learning complex patterns and handling long sequences. More recently, Transformer-based models~\cite{vaswani2017attention} have been adapted from NLP to time series~\cite{nietime, zhou2022fedformer, zhang2022self, eldele2021time, wu2021autoformer, zhou2021informer, koval-etal-2024-financial}, leveraging self-attention for long-range dependencies. 
Additionally, unsupervised representation learning methods pre-train models with masked time series modeling to minimize reconstruction error, followed by fine-tuning for downstream tasks like classification~\cite{franceschi2019unsupervised, tonekaboniunsupervised, yue2022ts2vec, eldele2021time, zerveas2021transformer, goswamimoment}. However, these complex models often require larger sizes and higher computational costs, particularly for training.

\paragraph{Language Models for Time Series}
The success of LLMs in NLP has inspired their application in time series analysis. Surveys~\cite{Zhang2024LargeLM, Jiang2024EmpoweringTS} provide insights into key methodologies, challenges, and future directions. Recent studies explore integrating time series and language, including structured time series-text modeling~\cite{khadanga-etal-2019-using, deznabi2021predicting}, natural language descriptions of time series~\cite{murakami2017learning, jhamtani-berg-kirkpatrick-2021-truth, fons-etal-2024-evaluating}, and various applications~\cite{li-etal-2024-econagent, drinkall-etal-2024-time, kawarada-etal-2024-demonstration}.  
Pre-trained LLMs have been used for time series forecasting via prompting~\cite{gruver2024large, liu-etal-2024-lstprompt, cao2023tempo}, while \cite{yu-etal-2023-harnessing} explored  their potential for generating explainable financial forecasts. Time-LLM~\cite{jin2023time} maps time series to the language embedding space, enabling LLM-based forecasting~\cite{yang2021voice2series}.  
More importantly, recent work, OneFitsAll~\cite{zhou2024one} achieved SOTA performance on various time series tasks by fine-tuning LLMs like GPT~\cite{radford2019language}. 
In contrast, we propose leveraging text embeddings instead of directly using LLMs for time series analysis. Our approach establishes new SOTA performance on time series classification tasks with significantly fewer trainable parameters, making it a more efficient alternative.

\paragraph{Text Embeddings}
Text embeddings are crucial in NLP, mapping words or sentences into dense vector spaces to capture semantic and syntactic information. Text embedding techniques range from word-level embeddings like Word2Vec~\cite{mikolov2013distributed} and GloVe~\cite{pennington2014glove} to contextualized embeddings from pre-trained models like BERT~\cite{devlin-etal-2019-bert} and RoBERTa~\cite{liu2019roberta}.  
In time series, unsupervised methods have been proposed for learning embeddings~\cite{franceschi2019unsupervised, eldele2021time, zerveas2021transformer, yue2022ts2vec, sun2023test, goswamimoment}. However, large-scale datasets are scarcer in time series than in NLP, making it more challenging to learn embeddings from scratch. To our knowledge, we are the \textbf{first} to leverage well-trained text embeddings from NLP for time series classification.
\section{Methodology}\label{sec:methods}
\begin{figure*}[htb] 
    \centering
    \includegraphics[scale = 0.5]{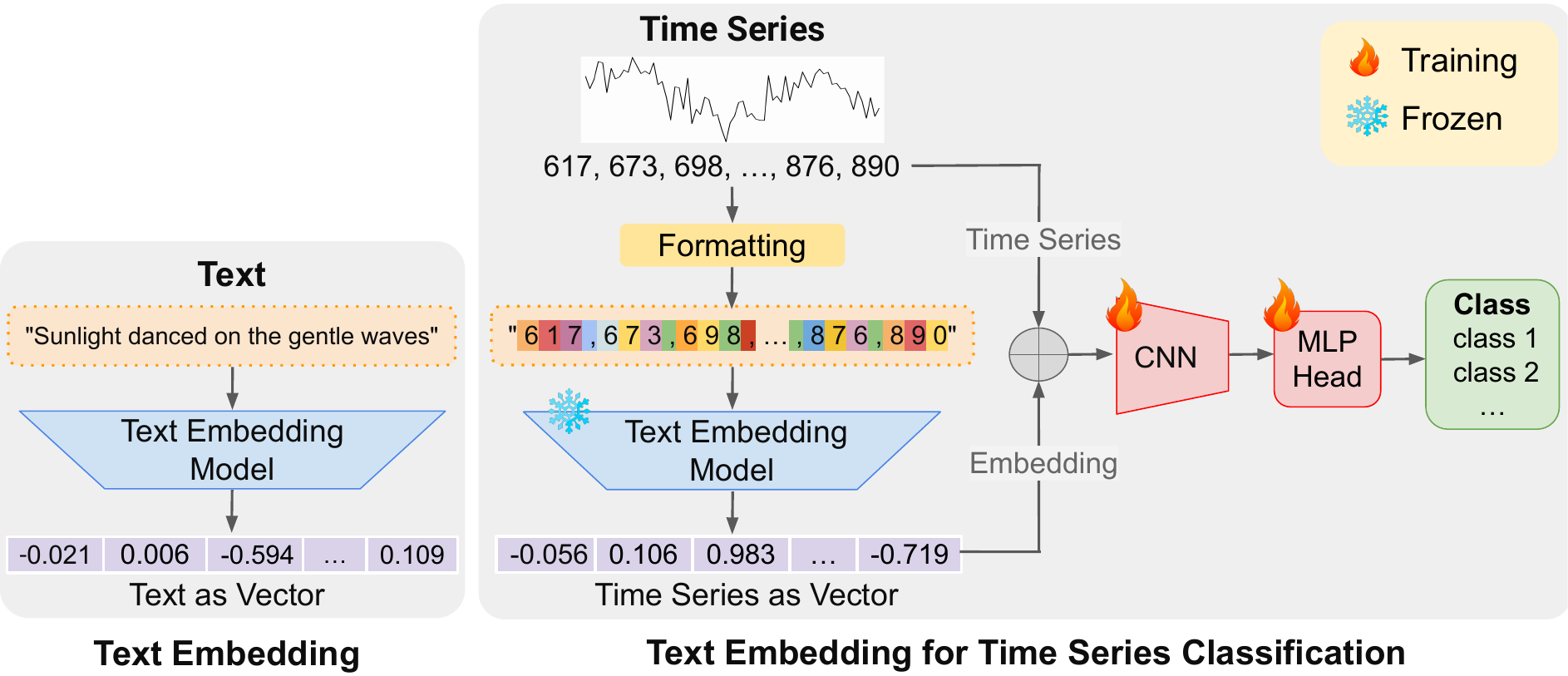}
    \caption{\textbf{Left:} Conventional text embedding. \textbf{Right:} Our proposed \letsc{} framework normalizes time series and formats them to tokenize each digit separately. It then embeds the time series, fuses the embeddings with the original series via element-wise addition, and uses a simple CNN-MLP classification head to perform classification. Only the lightweight CNN and MLP head are trained. For illustration, we show a single-dimensional time series $\mathbf{x_i} \in \mathbb{R}^{l_x}$, transformed into an embedding vector $\mathbf{e_i} \in \mathbb{R}^{l_e}$.
    } 
    \label{fig:methodology}
\end{figure*}

Given a time series classification dataset $\mathcal{D}=\{(\mathbf{x_i}, y_i)_{i=1}^N\}$, where $\mathbf{x_i}$ is a multivariate time series sample, and $y_i \in \{1, 2, \dots, C\}$ is the corresponding class label, the goal is to learn a classifier that accurately predicts the class label $\hat{y}_i$ for each time series.
As illustrated in Figure \ref{fig:methodology}, we propose \letsc{} framework that harnesses text embeddings for time series classification tasks. Specifically, we 1) initially preprocess the time series data to normalize it, then 2) subsequently generate text embeddings from the normalized time series, 3) fuse embeddings with the time series data, and finally 4) feed the fused representation to a classification head that consists of CNNs and MLP. The choice of a simple classification head is intentional, we aim to test the hypothesis that the text embeddings of the time series provide sufficiently powerful representations for effective classification.
\paragraph{Preprocessing}
To ensure consistent scales across all model inputs, each feature dimension of time series $\mathbf{x_i}$ is min-max normalized to the range $[0, 1]$ based on the minimum and maximum feature values of each dimension across the training data. 
\paragraph{Text Embedding of Time Series}
It is crucial to carefully format the preprocessed time series into strings before using text embeddings, as the tokenization of numerical strings can significantly affect the embeddings. ~\cite{liu2023goat} has shown that tokenization impacts a model's arithmetic abilities, with commonly used subword tokenization methods like Byte Pair Encoding (BPE) arbitrarily subdividing numbers, causing similar numbers to appear very differently. 
To mitigate this, we adopted a digit-space tokenization strategy, as suggested by~\cite{gruver2024large}, where each digit is spaced, commas are added to separate time steps, and decimal points are omitted for fixed precision. For instance, a series to be formatted with a precision of two decimal places, such as 0.645, 6.45, 64.5, 645.0, would be converted to "6 4 , 6 4 5 , 6 4 5 0 , 6 4 5 0 0" prior to tokenization. This method ensures separate tokenization of each digit, preserving numerical integrity and enhancing pattern recognition in language models.

Next, we utilized the \texttt{text-embedding-3-large} model \cite{openai_large_embeddings} to embed the formatted time series into the embedding space. It is important to note that we are using only the text embedding model, unlike the LLMs used in previous works \cite{jin2023time, zhou2024one, gruver2024large}.
This model was selected for several reasons: it is highly ranked on the Massive Text Embedding Benchmark (MTEB) leaderboard \cite{muennighoff2022mteb, MTEBleaderboard}, known for its effectiveness in a variety of downstream tasks such as text search and sentence similarity; it supports a high maximum token length of 8191, accommodating our time series datasets; and it offers a high-dimensional vector space of 3072 dimensions. This large dimensionality captures a broad spectrum of temporal features, additionally the model allows for truncation to reduce dimensions as needed for specific applications without substantial loss of semantic information \cite{kusupati2022matryoshka}. This capability to truncate dimensions is particularly advantageous for optimizing efficiency while maintaining robust performance, aligning well with our goal of a lightweight framework.

In particular, we generate a text embedding for each dimension of $\mathbf{x_i}$, transforming $\mathbf{x_i} \in \mathbb{R}^{d \times l_x}$ into embeddings $\mathbf{e_i} \in \mathbb{R}^{d \times l_e}$. Here, $d$ is the number of dimensions in the multivariate time series, $l_x$ is the series length, and $l_e$ is the embedding length. Specifically, each dimension of $\mathbf{x_i}$ (of length $l_x$) is first converted into a string, resulting in $d$ separate strings. 
Each string is then independently embedded using an embedding model, as studies \cite{nietime, zhou2024one, goswamimoment} suggest that channel-wise modeling is an effective strategy for multivariate time series.
Our approach differs slightly depending on the type of embedding model used:  
1) \textit{Proprietary model:} For OpenAI’s $\texttt{text-embedding-3-large}$, the API directly returns an embedding vector of size $l_e$ for each string;  
2) \textit{Open-weight models:} We extract the last token's embedding from the model's final hidden states to represent the entire sequence, effectively transforming the dimension from $l_x$ to $l_e$. This approach captures the final contextual representation by condensing the sequence into a single embedding derived from the last token. Embeddings for each dimension are then combined into a $d \times l_e$ matrix, effectively transforming the input from $d \times l_x$ to $d \times l_e$. This transformation preserves independent contextual representations for each dimension while ensuring a consistent embedding structure.
Note that the embedding computation in \letsc{} is a one-time pass, then the computed embeddings of the time series can be stored and reused for further training, in contrast to the persistent computational cost caused by fine-tuning parts of LLMs.
To validate the suitability of off-the-shelf text embeddings for time series classification, we compared cosine similarities between text embeddings of time series from the same and different classes. Intra-class similarity was consistently higher, demonstrating their effectiveness (see Section~\ref{effectiveness_3sections}).

\paragraph{Fusing Embedding and Time Series}
Next, we fuse the embedding with the preprocessed time series using element-wise addition, applying zero padding for dimensional consistency. The time series embedding vector serves as a feature representation of temporal patterns present in the entire time series, and adding the raw time series integrates both information sources. This approach is well-established, particularly in ResNet \cite{he2016deep}, where raw data is combined with feature representations via shortcut connections. Additionally, fusing embeddings from different modalities is widely supported \cite{manzoor2023multimodality, guo2019deep, poria2018multimodal, baltruvsaitis2018multimodal, jabri2016revisiting}, with element-wise addition being a common and effective method for combining multimodal embeddings. This fusion allows both time series and text embeddings to contribute meaningfully to the final representation, improving the model’s overall performance. 
As shown in Section \ref{sec:additional_results}, using either the text embedding or the raw time series alone, as well as alternative fusion methods to addition, is less effective. While element-wise addition is an empirical design choice in our architecture, our experiments demonstrate that it achieves SOTA performance while maintaining minimal model complexity compared to alternatives such as concatenation, or fusion.
\paragraph{Lightweight Classification Head}
Lastly, we pair the fused time series representation with a simple classification head composed of 1D CNNs and an MLP for time series classification. The output from CNNs are flattened and fed through the final MLP head with a softmax activation, which outputs a vector of the probabilities of each time series class. 
As detailed in the Appendix \ref{appendix:hyperparameter_details},
hyperparameter search determines the number of convolutional blocks in CNN,  the number of linear layers in MLP, and the use of batch normalization, dropout, activation, and pooling. 
With a simple classification head, our model is lightweight and requires much fewer trainable parameters compared to the existing SOTA built on transformers, as detailed in Section~\ref{results:performance_efficiency}.

\section{Experimental Protocol and Details}\label{sec:experiments}

\paragraph{Datasets and Evaluation Metrics}
We evaluated \letsc{} against a well-established benchmark for multivariate time series classification, commonly adopted for comparison in various studies \cite{zhou2024one, li2024time, wu2022timesnet, zerveas2021transformer}. This benchmark, selected from the UEA Archive \citep{bagnall2018uea}, is purposefully curated to cover challenging and diverse domains, including the following datasets: EthanolConcentration \citep{large2018detecting}, FaceDetection \citep{face_detection_dataset}, Handwriting \citep{shokoohi2017generalizing}, Heartbeat \citep{liu2016open}, JapaneseVowels \citep{misc_japanese_vowels_128}, PEMS-SF \citep{cuturi2011fast}, SelfRegulationSCP1 \citep{birbaumer1999spelling}, SelfRegulationSCP2 \citep{birbaumer1999spelling}, SpokenArabicDigits \citep{misc_spoken_arabic_digit_195}, and UWaveGestureLibrary \citep{liu2009uwave}. This commonly adopted benchmark offers a comprehensive testing environment, with multivariate dimensions ranging from 3 to 963, time series lengths up to 1751, and up to 26 classes.
See Appendix \ref{appendix:dataset_details} for details on each dataset.
To assess the classifiers, we used metrics including classification accuracy and \textit{AvgWins}. \textit{AvgWins} is defined as the average number of times that a method outperforms other methods across benchmarked datasets, with ties also being counted towards this average. Additionally, we analyzed models' computational efficiency in terms of trainable model parameters.
\paragraph{Baselines}
We included 27 baseline models to ensure a comprehensive comparison. We utilize the same supervised learning baselines outlined by \cite{zhou2024one, wu2022timesnet}, namely:
\noindent \textbf{Classical methods:} 1) Dynamic Time Warping (DTW) \citep{berndt1994using}, 2) eXtreme Gradient Boosting (XGBoost) \citep{chen2016xgboost}, and 3) RandOm Convolutional KErnel Transform (ROCKET) \citep{dempster2020rocket};
\noindent \textbf{MLP-based methods:} 4) LightTS \citep{zhang2022less}, and 5) DLinear \citep{zeng2023transformers};
\noindent \textbf{RNN-based models:} 6) Long Short-Term Memory (LSTM) \citep{hochreiter1997long}, 7) Long- and Short-term Time-series Network (LSTNet) \citep{lai2018modeling}, and 8) Linear State Space Layer (LSSL) \citep{gu2021efficiently};
\noindent \textbf{CNN-based models:} 9) Temporal Convolutional Network (TCN) \citep{franceschi2019unsupervised}, and 10) TimesNet \citep{wu2022timesnet};
\noindent \textbf{Transformer-based models:} 11) Transformer \citep{vaswani2017attention}, 12) Reformer \citep{kitaev2020reformer}, 13) Informer \citep{zhou2021informer}, 14) Pyraformer \citep{liu2021pyraformer}, 15) Autoformer \citep{wu2021autoformer}, 16) Non-stationary Transformer \citep{liu2022non}, 17) FEDformer \citep{zhou2022fedformer}, 18) ETSformer \citep{woo2022etsformer}, 19) Flowformer \citep{wu2022flowformer}, 20) Patch Time Series Transformer (PatchTST) \cite{nietime}; and \noindent \textbf{LLM-based model:} 21) OneFitsAll \citep{zhou2024one}. For details, see Appendix \ref{appendix:supervised_baselines}.
Note that some of these methods were adapted from forecasting to classification tasks, without altering the core design of the models (Appendix \ref{adapting_forecasting_to_classification}). 
Additionally, we included the unsupervised representation learning baselines outlined by \cite{goswamimoment}, namely: \noindent \textbf{CNN-based methods:} 22) T-Loss \cite{franceschi2019unsupervised}, 23) Temporal Neighborhood Coding (TNC) \cite{tonekaboniunsupervised}, 24) TS2Vec \cite{yue2022ts2vec}; \noindent \textbf{Transformer-based models:} 25) Time Series representation learning framework via Temporal and Contextual Contrasting (TS-TCC) \cite{eldele2021time}, 26) Time Series Transformer (TST) \cite{zerveas2021transformer}, and 27) MOMENT \cite{goswamimoment}. 
For details, refer to Appendices~\ref{appendix:unsupervised_baselines} (unsupervised baselines) and~\ref{appendix_adapting_unsupervised_to_classification} (adapting unsupervised models for classification). Reproduction details for baselines and LETS-C implementation details are in Appendices~\ref{appendix_reproduce_baselines} and~\ref{appendix:implementation_details}, respectively.

\section{Results and Analysis}\label{sec:results}
\subsection{Performance and Efficiency}\label{results:performance_efficiency}
\paragraph{\textit{Comparison to State-of-the-art}}
Table \ref{table:accuracy} and Figure \ref{fig:barplot_accuracy} present a comparative analysis of \letsc{} against 27 baseline models on the benchmark introduced above\footnote{To maintain consistency with prior works, we use the same settings as in TimesNet and OneFitsAll, where the UEA benchmark includes only training and testing sets (with no validation set). Consequently, the reported performance reflects the model's upper bound measurement. We collect supervised baseline model results from \cite{zhou2024one} and \cite{wu2022timesnet}, except for PatchTST, which is based on our own reproduction. Additionally, we gather unsupervised representation learning baseline results from \cite{goswamimoment} and \cite{yue2022ts2vec}.}.
\begin{table*}[htb]
\caption{Comparison of classification accuracy (\%) and AvgWins (\%).
\textcolor{red}{\textbf{Red:}} Best,
\textcolor{blue}{\underline{Blue:}} Second best. 
\textbf{Abbreviations:} EC: Ethanol Concentration, FD: Face Detection, HW: Handwriting, HB: Heartbeat, JV: Japanese Vowels, SCP1: Self-Regulation SCP1, SCP2: Self-Regulation SCP2, SAD: Spoken Arabic Digits, UW: UWave Gesture Library.}
\label{table:accuracy}
\centering
\resizebox{\linewidth}{!}{%
\begin{tabular}{@{}p{1.9cm}p{6.8cm}|cccccccccc|c|c@{}}
\toprule
\multicolumn{2}{c|}{Model/Dataset} & EC & FD & 
HW & 
HB & JV & PEMS-SF & SCP1 & SCP2 & SAD & UW & \cellcolor{pale7} \textbf{Average} $\mathbf{\uparrow}$ & \cellcolor{pale7} \textbf{AvgWins \%} $\mathbf{\uparrow}$ \\ \midrule
\multicolumn{14}{l}{\cellcolor{pale4}\textit{Supervised Learning Methods}}
\\ 
\midrule
\multirow{3}{1.9cm}{Classical methods} 
& DTW \cite{berndt1994using} & 32.3 & 52.9 & 
28.6 & 
71.7 & 94.9 & 71.1 & 77.7 & 53.9 & 96.3 & 90.3 & 
\cellcolor{pale7}  66.97 & \cellcolor{pale7} 0\%
\\
& XGBoost  \cite{chen2016xgboost} & 43.7    & 63.3 
& 15.8 
& 73.2 & 86.5 & 98.3 & 84.6 & 48.9 & 69.6 & 75.9 & 
\cellcolor{pale7} 65.98 & \cellcolor{pale7} 10\%
\\
& ROCKET \cite{dempster2020rocket} & 45.2    & 64.7 
& 58.8 
& 75.6 & 96.2 & 75.1 & 90.8 & 53.3 & 71.2 & 94.4 & 
\cellcolor{pale7} 72.53 & \cellcolor{pale7} \textcolor{blue}{\underline{20\%}}
\\
\midrule
\multirow{2}{*}{MLP} 
& LightTS \cite{zhang2022less} & 29.7    & 67.5 
& 26.1 
& 75.1 & 96.2 & 88.4 & 89.8 & 51.1 & 100  & 80.3 & 
\cellcolor{pale7} 70.42 & \cellcolor{pale7} 10\%
\\
& DLinear \cite{zeng2023transformers} & 32.6    & 68   
& 27   
& 75.1 & 96.2 & 75.1 & 87.3 & 50.5 & 81.4 & 82.1 & 
\cellcolor{pale7} 67.53 & \cellcolor{pale7}  0\%
\\
\midrule
\multirow{4}{*}{RNN} 
& LSTM \cite{hochreiter1997long} & 32.3 & 57.7 
& 15.2 
& 72.2 & 79.7 & 39.9 & 68.9 & 46.6 & 31.9 & 41.2 & 
\cellcolor{pale7} 48.56 & \cellcolor{pale7} 0\%
\\
& LSTNet \cite{lai2018modeling} & 39.9    & 65.7 
& 25.8 
& 77.1 & 98.1 & 86.7 & 84   & 52.8 & 100  & 87.8 & 
\cellcolor{pale7} 71.79 & \cellcolor{pale7} 10\%
\\
& LSSL  \cite{gu2021efficiently} & 31.1    & 66.7 
& 24.6 
& 72.7 & 98.4 & 86.1 & 90.8 & 52.2 & 100  & 85.9 & 
\cellcolor{pale7} 70.85 & \cellcolor{pale7} 10\%
\\
\midrule
\multirow{2}{*}{CNN} 
&  TCN \cite{franceschi2019unsupervised}  & 28.9    & 52.8 
& 53.3 
& 75.6 & 98.9 & 68.8 & 84.6 & 55.6 & 95.6 & 88.4 & 
\cellcolor{pale7} 70.25 & \cellcolor{pale7} 0\%
\\
& TimesNet \cite{wu2022timesnet} & 35.7    & 68.6 
& 32.1 
& 78   & 98.4 & 89.6 & 91.8 & 57.2 & 99   & 85.3 & 
\cellcolor{pale7} 73.57 & \cellcolor{pale7} 0\%
\\
\midrule
\multirow{9}{*}{Transformer} & Transformer \cite{vaswani2017attention} & 32.7    & 67.3 
& 32   
& 76.1 & 98.7 & 82.1 & 92.2 & 53.9 & 98.4 & 85.6 & 
\cellcolor{pale7} 71.9  & \cellcolor{pale7} 0\%
\\
&  Reformer \cite{kitaev2020reformer} & 31.9    & 68.6 
& 27.4 
& 77.1 & 97.8 & 82.7 & 90.4 & 56.7 & 97   & 85.6 & 
\cellcolor{pale7} 71.52 & \cellcolor{pale7} 0\%
\\
&  Informer \cite{zhou2021informer} & 31.6    & 67   
& 32.8 
& 80.5 & 98.9 & 81.5 & 90.1 & 53.3 & 100  & 85.6 & 
\cellcolor{pale7} 72.13 & \cellcolor{pale7} \textcolor{blue}{\underline{20\%}}
\\
& Pyraformer \cite{liu2021pyraformer} & 30.8    & 65.7 
& 29.4 
& 75.6 & 98.4 & 83.2 & 88.1 & 53.3 & 99.6 & 83.4 & 
\cellcolor{pale7} 70.75 & \cellcolor{pale7} 0\%
\\
& Autoformer \cite{wu2021autoformer} & 31.6    & 68.4 
& 36.7 
& 74.6 & 96.2 & 82.7 & 84   & 50.6 & 100  & 85.9 & 
\cellcolor{pale7} 71.07 & \cellcolor{pale7} 10\%
\\
& Non-stationary Transformer \cite{liu2022non} & 32.7    & 68   
& 31.6 
& 73.7 & 99.2 & 87.3 & 89.4 & 57.2 & 100  & 87.5 & 
\cellcolor{pale7} 72.66 & \cellcolor{pale7} \textcolor{blue}{\underline{20\%}}
\\
& FEDformer \cite{zhou2022fedformer} & 31.2    & 66   
& 28   
& 73.7 & 98.4 & 80.9 & 88.7 & 54.4 & 100  & 85.3 & 
\cellcolor{pale7} 70.66 & \cellcolor{pale7} 10\%
\\
& ETSformer \cite{woo2022etsformer} & 28.1    & 66.3 
& 32.5 
& 71.2 & 95.9 & 86   & 89.6 & 55   & 100  & 85   & 
\cellcolor{pale7} 70.96 & \cellcolor{pale7} 10\%
\\
& Flowformer \cite{wu2022flowformer} & 33.8    & 67.6 
& 33.8 
& 77.6 & 98.9 & 83.8 & 92.5 & 56.1 & 98.8 & 86.6 & 
\cellcolor{pale7} 72.95 & \cellcolor{pale7} 0\%
\\
& PatchTST \cite{nietime} & 26.2 & 68.5 & 25.9 & 66.8 & 96.0 & 87.9 & 85.7 & 53.3 & 97.2 & 85.0 & \cellcolor{pale7} 69.25 &  \cellcolor{pale7} 0\%
\\
\midrule
\multirow{1}{*}{LLM} & OneFitsAll \cite{zhou2024one} & 34.2    & 69.2 
& 32.7 
& 77.2 & 98.6 & 87.9 & 93.2 & 59.4 & 99.2 & 88.1 & 
\cellcolor{pale7} \textcolor{blue}{\underline{73.97}} & \cellcolor{pale7} \textcolor{blue}{\underline{20\%}}
\\
\midrule
\midrule
\multicolumn{14}{l}{\cellcolor{pale4}\textit{Unsupervised Representation Learning Methods}}
\\ 
\midrule
\multirow{3}{*}{CNN} 
& T-Loss \cite{franceschi2019unsupervised} & 20.5 & 51.3
& 45.1
& 74.1 & 98.9 &  67.6 & 84.3 & 53.9 & 90.5 & 87.5 & 
\cellcolor{pale7} 67.37 & \cellcolor{pale7} 0\%
\\
& TNC \cite{tonekaboniunsupervised} & 29.7 & 53.6
& 24.9
& 74.6 & 97.8 &  69.9 & 79.9 & 55.0 & 93.4 & 75.9 & 
\cellcolor{pale7} 65.47 & \cellcolor{pale7} 0\%
\\
& TS2Vec \cite{yue2022ts2vec} & 30.8 & 50.1
& 51.5
& 68.3 & 98.4 &  68.2 & 81.2 & 57.8 & 98.8 & 90.6 & 
\cellcolor{pale7} 69.57 & \cellcolor{pale7} 0\%
\\
\midrule
\multirow{3}{*}{Transformer} 
& TS-TCC \cite{eldele2021time}  & 28.5 & 54.4
& 49.8
& 75.1 & 93.0 &  73.4 & 82.3 & 53.3 & 97.0 & 75.3 & 
\cellcolor{pale7} 68.21 & \cellcolor{pale7} 0\%
\\
& TST \cite{zerveas2021transformer}  & 26.2  & 53.4
& 22.5
& 74.6 & 97.8 &  74.0 & 75.4 & 55.0 & 92.3 & 57.5 & 
\cellcolor{pale7} 62.87 & \cellcolor{pale7} 0\%
\\
& MOMENT \cite{goswamimoment} & 35.7 & 63.3
& 30.8 
& 72.2 & 71.6 & 89.6 & 84.0 & 47.8 & 98.1 & 90.9 & 
\cellcolor{pale7} 68.4 & \cellcolor{pale7} 0\%
\\
\midrule
\midrule
& \textbf{\letsc{} (Ours)} & 52.9 & 68.9
& 23.8
& 78   & 99.2 & 93.1 & 93.2 & 62.8 & 99.2 & 90.6 & 
\cellcolor{pale7} \textcolor{red}{\textbf{76.17}} & \cellcolor{pale7} \textcolor{red}{\textbf{40\%}}
\\ 
\bottomrule
\end{tabular}
}
\end{table*}
\begin{table*}[htb]
\caption{Comparison of trainable parameters (millions) for \letsc{} vs. OneFitsAll. 
The Ratio (\%) \(= 100 \times \frac{\text{\# of trainable parameters in \letsc{}}}{\text{\# of trainable parameters in OneFitsAll}}\), quantifying the efficiency of \letsc{} relative to OneFitsAll.}
\label{table:trainable_parameters}
\centering
\resizebox{\linewidth}{!}{%
\begin{tabular}{@{}p{2.2cm}l|cccccccccc|c@{}}
\toprule
& Model/Dataset & EC & FD 
& HW 
& HB & JV & PEMS-SF & SCP1 & SCP2 & SAD & UW & \cellcolor{pale7} {\textbf{Average} $\mathbf{\downarrow}$} \\ \midrule
\multirow{3}{2.2cm}{Trainable parameters (M)} 
& \letsc{} (Ours) & 0.28 & 0.003 & 0.15 & 0.04 & 0.14 & 0.56 & 0.30 & 0.33 & 0.14 & 0.26 & \cellcolor{pale7} 0.22 \\
& OneFitsAll    & 1.42 & 2.37 & 1.73 & 2.03 & 1.32 & 10.23 & 0.98   & 1.04 & 1.82  & 1.0 & \cellcolor{pale7} 2.39 \\
\cmidrule{2-13}
& \cellcolor{pale7} \textbf{Ratio (\%)} & \cellcolor{pale7} 19.89 & \cellcolor{pale7} 0.16 & \cellcolor{pale7} 8.89 & \cellcolor{pale7} 2.28 & \cellcolor{pale7} 11.19 & \cellcolor{pale7} 5.51 & \cellcolor{pale7} 30.83 & \cellcolor{pale7} 32.06 & \cellcolor{pale7} 7.77 & \cellcolor{pale7} 26.21 & \cellcolor{pale7} \textcolor{magenta}{\textbf{14.48}} \\
\bottomrule
\end{tabular}
}
\end{table*}
We observe that \letsc{} consistently demonstrates robust performance across all datasets, achieving the highest average accuracy of 76.16\% and AvgWins of 40\%, compared to 27 benchmark models. This includes the most recent SOTA model OneFitsAll (accuracy: 73.97\%, AvgWins: 20\%) and an older SOTA TimesNet (accuracy: 73.57\%, AvgWins: 0\%). Notably, \letsc{} surpasses OneFitsAll on six out of ten datasets by a significant margin. 
\letsc{} is particularly effective on challenging datasets like PEMS-SF and EthanolConcentration (EC). PEMS-SF is characterized by exceptionally high dimensionality with 963 features, and EC contains an extremely long time series at length of 1751.
These results showcase the competitive edge of \letsc{} against the previous SOTA methods, thus establishing a new benchmark for time series classification.

\paragraph{\textit{Computational Efficiency}}
Next, we aim to assess how well \letsc{} balances performance with computational efficiency, which is crucial for usage in resource-constrained environments. 
Table \ref{table:trainable_parameters} provides a detailed analysis of the trainable parameters associated with \letsc{} compared to the previous SOTA model, OneFitsAll. 
Our method achieved higher performance with only 14.48\% of the trainable model parameters on average, compared to OneFitsAll. Despite the advantage of OneFitsAll over other leading models like TimesNet and FEDformer on its reduced parameter count, OneFitsAll still requires much more trainable parameters than our approach.
We show that \letsc{} offers a lightweight approach to  time series classification while achieving the SOTA accuracy.
It’s important to note that in \letsc{}, the text embedding computation is a one-time operation, unlike models like OneFitsAll, which continue to incur computational costs while fine-tuning partially frozen LLMs.
\subsection{Effectiveness of \letsc{}} \label{effectiveness_3sections}
\begin{figure*}[htb]
    \centering
    \includegraphics[scale = 0.55]{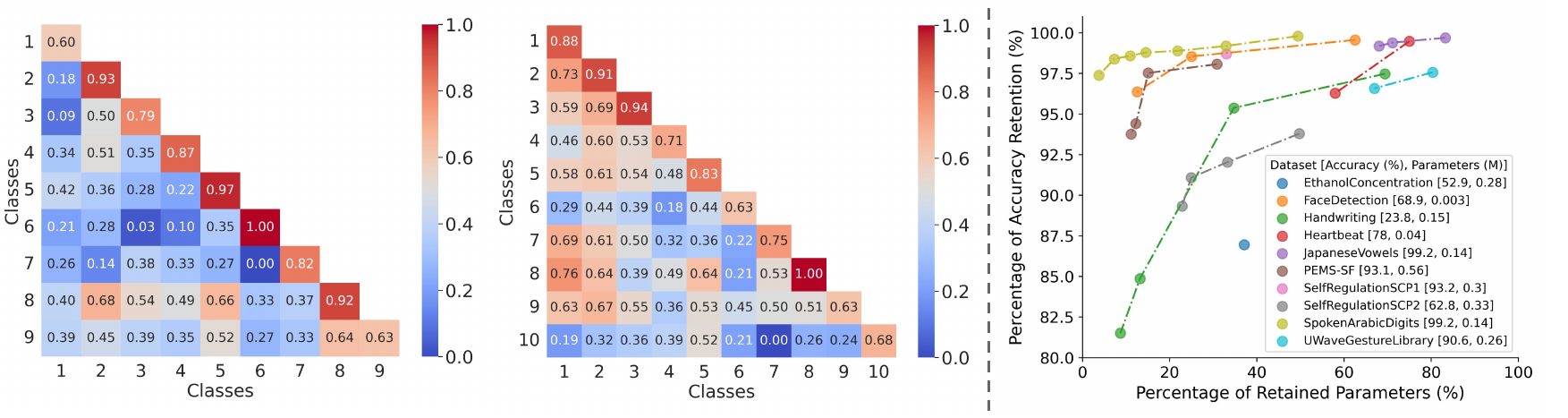}
    \caption{\textbf{Left:} Heatmaps illustrating within- and between-class cosine similarities of text embeddings derived from the training time series in JapaneseVowels \textbf{(far left)} with 9 classes and SpokenArabicDigits \textbf{(middle)} with 10 classes. 
    \textbf{Right:} Trade-off between the percentage of accuracy retention and model parameter retention relative to \letsc{}'s optimal values. The optimal \letsc{} accuracy (\%) and parameters (M) for each dataset are  in the legend. 
    }
\label{cosine_similarity_train_tradeoff_combined}
\end{figure*}

\paragraph{\textit{Built-in Discriminative Power of Text Embeddings on Time Series}}
To analyze text embeddings extracted from time series and assess their effectiveness,
we compared embeddings of time series from the same and different classes. This study revealed the built-in power of text embeddings to readily distinguish time series from different classes.
Specifically, we computed the average cosine similarity for time series pairs within the same class and across different classes.  
We average similarities from multiple channels to handle multivariate time series effectively. This approach helps us quantify how closely time series in the embedding space are related, both within each class (intra-class) and across distinct classes (inter-class). 
Figure \ref{cosine_similarity_train_tradeoff_combined} (left) visualizes these embedding similarities through heatmaps. The heatmaps are scaled using min-max normalization, 
where warmer colors in the heatmap represent higher similarities and cooler shades indicate lower similarities. Diagonal entries show intra-class similarities, while off-diagonal entries reveal inter-class similarities.
\begin{table*}[htb]
\caption{
Comparison of \letsc{} with various embedding models against OneFitsAll. Performances surpassing OneFitsAll are in \textbf{bold}, with the best in \textcolor{red}{\textbf{Red}}. AvgWins scores above 50\% indicate consistent superiority, calculated as 1 for outperforming OneFitsAll and 0 otherwise. Ratio (\%) \(=100 \times \frac{\text{\# of trainable parameters in \letsc{}}}{\text{\# of trainable parameters in OneFitsAll}}\) measures computational efficiency relative to OneFitsAll.
}
\label{table:embedding_type_performance}
\centering
\resizebox{\linewidth}{!}{%
\begin{tabular}{@{}ll|cccccccccc|c|c@{}}
\toprule
\multicolumn{14}{c}{\bf Accuracy $\mathbf{\uparrow}$} \\
\midrule
\multicolumn{2}{l|}{Method/Dataset} & EC & FD & 
HW & 
HB & JV & PEMS-SF & SCP1 & SCP2 & SAD & UW 
& 
\cellcolor{pale7} \textbf{Average} 
$\mathbf{\uparrow}$  
& 
\cellcolor{pale7} \textbf{AvgWins \%} $\mathbf{\uparrow}$ 
\\ 
\midrule
\multicolumn{2}{l|}{OneFitsAll } & 34.2    & 69.2 & 32.7 & 77.2 & 98.6 & 87.9 & 93.2 & 59.4 & 99.2 & 88.1 & \cellcolor{pale7} 73.97 & \cellcolor{pale7} $-$ 
\\
\midrule
\; \multirow{4}{1.4cm}{\letsc{}} & \texttt{text-embedding-3-large} & 52.9 & 68.9
& 23.8 & 78   & 99.2 & 93.1 & 93.2 & 62.8 & 99.2 & 90.6 & \cellcolor{pale7} \textcolor{red}{\textbf{76.17}} & \cellcolor{pale7} \textcolor{red}{\textbf{80\%}} \\
\cmidrule{2-14}
& \texttt{e5-mistral-7b-instruct} & 55.5 & 68.7 & 23.3 & 77.6 & 99.2 & 84.4 & 93.9 & 59.4 & 98.5 & 88.4 & \cellcolor{pale7} \textbf{74.89} & \cellcolor{pale7} \textbf{60\%} \\
\cmidrule{2-14}
& \texttt{gte-large-en-v1.5} & 57.8 & 68.8 & 24.7 & 77.6 & 98.4 & 91.3 & 94.2 & 60 & 99 & 88.4 & \cellcolor{pale7} \textbf{76.02} & \cellcolor{pale7} \textbf{60\%} \\
\cmidrule{2-14}
& \texttt{nomic-embed-text-v1} & 52.9 & 68 &  24.8 & 76.6 & 99.2 & 88.4 & 93.9 & 59.4 & 98.6 & 89.4 & \cellcolor{pale7} \textbf{75.12} & \cellcolor{pale7} \textbf{60\%} \\
\bottomrule
\toprule
\multicolumn{14}{c}{\bf Trainable Parameters (M) $\downarrow$} \\
\midrule
\multicolumn{2}{l|}{Method/Dataset} & EC & FD & 
HW & 
HB & JV & PEMS-SF & SCP1 & SCP2 & SAD & UW 
& 
\multicolumn{2}{c}{\cellcolor{pale7} \textbf{Average} $\mathbf{\downarrow}$} 
\\ 
\midrule
\multicolumn{2}{l}{OneFitsAll} & 1.42 & 2.37 & 1.73 & 2.03 & 1.32 & 10.23 & 0.98   & 1.04 & 1.82  & 1.0 & \multicolumn{2}{c}{\cellcolor{pale7} 2.39 }
\\
\midrule
\; \multirow{8}{1.4cm}{\letsc{}} & \texttt{text-embedding-3-large} & 0.28 & 0.003 & 0.15 & 0.04 & 0.14 & 0.56 & 0.30 & 0.33 & 0.14 & 0.26 & \multicolumn{2}{c}{\cellcolor{pale7} \textbf{0.22} } \\
& \quad \quad \quad \quad \quad \quad \quad \quad \quad  \quad \cellcolor{pale7} Ratio \% & \cellcolor{pale7} 19.89 & \cellcolor{pale7} 0.16 & \cellcolor{pale7} 8.89 & \cellcolor{pale7} 2.28 & \cellcolor{pale7} 11.19 & \cellcolor{pale7} 5.51 & \cellcolor{pale7} 30.83 & \cellcolor{pale7} 32.06 & \cellcolor{pale7} 7.77 & \cellcolor{pale7} 26.21 & \multicolumn{2}{c}{\cellcolor{pale7} \textcolor{magenta}{\textbf{14.48}}} \\
\cmidrule{2-14}
& \texttt{e5-mistral-7b-instruct} & 0.13 & 0.40 & 0.39 & 0.17 & 0.16 & 0.30 & 0.24 & 0.24 & 0.33 & 0.16 & \multicolumn{2}{c}{\cellcolor{pale7} \textbf{0.25} } \\
& \quad \quad \quad \quad \quad \quad \quad \quad \quad  \quad \cellcolor{pale7} Ratio \% & \cellcolor{pale7} 9.48 & \cellcolor{pale7} 16.93 & \cellcolor{pale7} 22.78 & \cellcolor{pale7} 8.54 & \cellcolor{pale7} 12.55 & \cellcolor{pale7} 2.97 & \cellcolor{pale7} 25.06 & \cellcolor{pale7} 23.59 & \cellcolor{pale7} 18.39 & \cellcolor{pale7} 15.93 & \multicolumn{2}{c}{\cellcolor{pale7} \textcolor{magenta}{\textbf{15.62}}} \\
\cmidrule{2-14}
& \texttt{gte-large-en-v1.5} & 0.18 & 0.31 & 0.31 & 0.12 & 0.04 & 0.56 & 0.03 & 0.07 & 0.22 & 0.09 & \multicolumn{2}{c}{\cellcolor{pale7} \textbf{0.19} } \\
& \quad \quad \quad \quad \quad \quad \quad \quad \quad  \quad 
 \cellcolor{pale7} Ratio \% & \cellcolor{pale7} 12.91 & \cellcolor{pale7} 13.44 & \cellcolor{pale7} 18.27 & \cellcolor{pale7} 6.11 & \cellcolor{pale7} 3.36 & \cellcolor{pale7} 5.51 & \cellcolor{pale7} 3.77 & \cellcolor{pale7} 7.65 & \cellcolor{pale7} 12.34 & \cellcolor{pale7} 9.00 & \multicolumn{2}{c}{\cellcolor{pale7} \textcolor{magenta}{\textbf{9.24}}} \\
\cmidrule{2-14}
& \texttt{nomic-embed-text-v1} & 0.03 & 0.03 & 0.36 & 0.07 & 0.05 & 0.19 & 0.02 & 0.10 & 0.06 & 0.02 & \multicolumn{2}{c}{\cellcolor{pale7} \textcolor{red}{\textbf{0.09}} } \\
& \quad \quad \quad \quad \quad \quad \quad \quad \quad  \quad 
   \cellcolor{pale7} Ratio \% & \cellcolor{pale7} 2.21 & \cellcolor{pale7} 1.31 &  \cellcolor{pale7} 20.99 &  \cellcolor{pale7} 3.58 &  \cellcolor{pale7} 3.99 &  \cellcolor{pale7} 1.85 &  \cellcolor{pale7} 3.02 &  \cellcolor{pale7} 10.03 &  \cellcolor{pale7} 3.34 &  \cellcolor{pale7} 2.78 &  \multicolumn{2}{c}{\cellcolor{pale7} \textcolor{magenta}{\textbf{5.31}}} \\
\bottomrule
\end{tabular}
}
\end{table*}
We observe that intra-class similarity consistently exceeds inter-class similarity, demonstrating that text embeddings effectively retain and convey significant information from the underlying time series. 

\paragraph{\textit{Generalization Across Various Text Embedding Models}}
To assess the generalization of our approach across various text embedding models beyond \texttt{text-embedding-3-large}, we evaluate \letsc{} using three additional embeddings: \texttt{e5-mistral-7b-instruct} \cite{wang-etal-2024-improving-text}, \texttt{gte-large-en-v1.5} \cite{li2023towards}, and \texttt{nomic-embed-text-v1} \cite{nussbaum2024nomic}.
These models were selected for their high rankings in the MTEB overall leaderboard and their variations in embedding dimensions, maximum token lengths, and model sizes.
Details on embedding models are in Appendix \ref{appendix:emb_type_analysis}.
Table \ref{table:embedding_type_performance} presents detailed accuracy metrics and trainable parameters for these various embedding models in the \letsc{} framework.
We observe that \letsc{} consistently outperforms current baselines in terms of average classification accuracy and AvgWins, while utilizing only a fraction of the trainable model parameters across all explored text embedding models
Specifically, \texttt{text-embedding-3-large} achieves an average accuracy of 76.17\%, while \texttt{e5-mistral-7b-instruct}, \texttt{gte-large-en-v1.5}, \& \texttt{nomic-embed-text-v1} achieve accuracies of 74.89\%, 76.02\%, and 75.12\% respectively. These results not only surpass the previous SOTA accuracy of 73.97\% but also require significantly fewer trainable model parameters—just 14.48\%, 15.62\%, 9.24\%, and 5.31\% compared to OneFitsAll. 
Among these text embedding variations, higher text embedding dimensionality leads to more trainable parameters on average. \texttt{e5-mistral-7b-instruct} requires the most trainable parameters (0.25M) due to its 4096 dimensions, followed by \texttt{text-embedding-3-large} (3072D, 0.22M), \texttt{gte-large-en-v1.5} (1024D, 0.19M), and \texttt{nomic-embed-text-v1} (768D, 0.09M).
Consequently, our approach generalizes across diverse text embedding models, demonstrates superior performance, with the benefit of being lightweight.

\paragraph{\textit{Optimizing Model Size with Minimal Accuracy Loss}}
Next, we examine the trade-offs between model accuracy and model size in our approach. To vary model size, we adjust the number of linear and convolution layers in the classification head, ranging from 1 to 5 layers each, creating model variants of different sizes. Across all datasets, we find that significant reductions in model parameters result in only a minimal loss of accuracy.  
Figure \ref{cosine_similarity_train_tradeoff_combined} (right) illustrates this trade-off, showing accuracy and parameter retention relative to the optimal performance of \letsc{} (see Tables \ref{table:accuracy} and \ref{table:trainable_parameters} for reference optimal values). 
The results highlight the efficiency of \letsc{}, demonstrating its ability to maintain high accuracy with significantly fewer parameters across all datasets. While the trade-off between model size and accuracy is data-dependent, in general, substantial parameter reductions lead to only slight accuracy decreases.  
For detailed results, see Appendix \ref{appendix:tradeoff} and Table \ref{appendixtable:tradeoff_accuracy_parameters}. This analysis confirms that while reducing parameters can lower accuracy, the impact remains manageable, making \letsc{} suitable for applications requiring efficient inference.
\subsection{Additional Analysis}\label{sec:additional_results}
\paragraph{\textit{Ablation Study}}
To empirically evaluate the benefits of fusing text embeddings with time series over variants using only one modality, we conduct an ablation study.
As shown in Appendix  \ref{appendix:ablation_study} and Table~\ref{table:ablation_embeddings}, we observe that the combination of both embeddings and time series achieves the highest average accuracy, at 76.17\%, and the best number of AvgWins, compared to the ablated versions. 
These demonstrate the significant performance gains from fusing embeddings with time series data, which are essential for optimal model accuracy.

\paragraph{\textit{Alternative Methods for Fusing Time Series with Embeddings}}
We explore two additional methods for fusing time series and embeddings beyond simple addition. The first method involves a \textit{Fusion network} that first processes embeddings and time series data through convolutional and dense layers in two separate branches, then merges the features from both branches into a final dense network. The second method employs \textit{Concatenation}, where the time series and embeddings are concatenated and processed through a lightweight classification head. Despite cross-attention being another alternative for fusing different modalities, we didn't include it in this study due to the computational complexity it adds to the model.
Appendix  \ref{appendix:alternative_fusion_methods} and Table \ref{table:alternative_methods_integrating} present details on classification accuracy and trainable model parameters for these variations.
We observe that the addition approach in the \letsc{} architecture achieves the highest average classification accuracy (76.11\%) compared to the fusion network (73.40\%) and concatenation (74.22\%). It also records the best AvgWins at 70\%. Further, the number of parameters increases with both the concatenation and fusion network approaches, resulting in additional complexity compared to the addition method. 
As our goal was to develop a lightweight model, we opted for the addition approach due to its simpler parameter structure.
\section{Conclusion}\label{sec:conclusion_future_work}
We introduced \letsc{}, a novel approach that leverages text embeddings for time series classification. To our knowledge, this is the first exploration of using off-the-shelf text embeddings in time series analysis, particularly for classification. By projecting time series  through text embedding models and employing a simple yet effective classification head, \letsc{} achieves SOTA performance on a well-established benchmark covering various domains, surpassing 27 baselines. Additionally, \letsc{} is significantly more lightweight than previous SOTA methods, achieving higher accuracy with fewer trainable parameters. Our comprehensive analysis highlights \letsc{}'s robustness across different text embedding models, the advantage of using text embeddings for time series classification, and the trade-off between accuracy and model size. 

\section{Limitations and Broader Perspective} \label{sec:limitations} 
\textbf{\textit{Limitations and future work:}}
Our approach has a few key limitations. One is the maximum sequence length constraint of text embedding models, which restricts the handling of longer time series and may necessitate truncation or downsampling, potentially affecting the capture of long-term dependencies. Additionally, while we adhered to established benchmarks for consistency, we did not evaluate our method on a newly constructed or strictly held-out dataset to check for potential data leakage into the OpenAI embeddings. Testing on a novel dataset could further validate our results and ensure that the approach’s effectiveness is not influenced by pre-existing knowledge. Another limitation is the monetary cost associated with using OpenAI's text embeddings, which we discuss in more detail in Appendix \ref{sec:monetary_costs}.

Future research could explore alternative time series tokenization strategies, as well as extensions to other time series tasks. We believe our findings will inspire further exploration of text embeddings in the time series domain, paving the way for more powerful and efficient methods for various time series tasks.

\textbf{\textit{Broader perspective:}} 
The embeddings generated may lack interpretability, making it difficult to understand model decisions in high-stakes applications. Additionally, the resources accompanying this study will be responsibly released for research purposes only.

\textbf{\textit{Datasets:}} 
The benchmarks used in this study are publicly available from the UEA Archive \citep{bagnall2018uea} and were curated by previous research. Specifically, they include the following datasets: EthanolConcentration \citep{large2018detecting}, FaceDetection \citep{face_detection_dataset}, Handwriting \citep{shokoohi2017generalizing}, Heartbeat \citep{liu2016open}, JapaneseVowels \citep{misc_japanese_vowels_128}, PEMS-SF \citep{cuturi2011fast}, SelfRegulationSCP1 \citep{birbaumer1999spelling}, SelfRegulationSCP2 \citep{birbaumer1999spelling}, SpokenArabicDigits \citep{misc_spoken_arabic_digit_195}, and UWaveGestureLibrary \citep{liu2009uwave}. We abide by their terms of use. 

 
\section*{Acknowledgments}
This paper was prepared for informational purposes by the Artificial Intelligence Research group of JPMorgan Chase \& Co and its affiliates (“J.P. Morgan”) and is not a product of the Research Department of J.P. Morgan.  J.P. Morgan makes no representation and warranty whatsoever and disclaims all liability, for the completeness, accuracy or reliability of the information contained herein.  This document is not intended as investment research or investment advice, or a recommendation, offer or solicitation for the purchase or sale of any security, financial instrument, financial product or service, or to be used in any way for evaluating the merits of participating in any transaction, and shall not constitute a solicitation under any jurisdiction or to any person, if such solicitation under such jurisdiction or to such person would be unlawful. 

\bibliography{custom}

\begin{thebibliography}{113}
\providecommand{\natexlab}[1]{#1}

\bibitem[{Abanda et~al.(2019)Abanda, Mori, and Lozano}]{abanda2019review}
Amaia Abanda, Usue Mori, and Jose~A Lozano. 2019.
\newblock A review on distance based time series classification.
\newblock \emph{Data Mining and Knowledge Discovery}, 33(2):378--412.

\bibitem[{Achiam et~al.(2023)Achiam, Adler, Agarwal, Ahmad, Akkaya, Aleman, Almeida, Altenschmidt, Altman, Anadkat et~al.}]{achiam2023gpt}
Josh Achiam, Steven Adler, Sandhini Agarwal, Lama Ahmad, Ilge Akkaya, Florencia~Leoni Aleman, Diogo Almeida, Janko Altenschmidt, Sam Altman, Shyamal Anadkat, et~al. 2023.
\newblock Gpt-4 technical report.
\newblock \emph{arXiv preprint arXiv:2303.08774}.

\bibitem[{Bagnall et~al.(2018)Bagnall, Dau, Lines, Flynn, Large, Bostrom, Southam, and Keogh}]{bagnall2018uea}
Anthony Bagnall, Hoang~Anh Dau, Jason Lines, Michael Flynn, James Large, Aaron Bostrom, Paul Southam, and Eamonn Keogh. 2018.
\newblock The uea multivariate time series classification archive, 2018.
\newblock \emph{arXiv preprint arXiv:1811.00075}.

\bibitem[{Bagnall et~al.(2017)Bagnall, Lines, Bostrom, Large, and Keogh}]{bagnall2017great}
Anthony Bagnall, Jason Lines, Aaron Bostrom, James Large, and Eamonn Keogh. 2017.
\newblock The great time series classification bake off: a review and experimental evaluation of recent algorithmic advances.
\newblock \emph{Data mining and knowledge discovery}, 31:606--660.

\bibitem[{Bahlmann et~al.(2002)Bahlmann, Haasdonk, and Burkhardt}]{bahlmann2002online}
Claus Bahlmann, Bernard Haasdonk, and Hans Burkhardt. 2002.
\newblock Online handwriting recognition with support vector machines-a kernel approach.
\newblock In \emph{Proceedings eighth international workshop on frontiers in handwriting recognition}, pages 49--54. IEEE.

\bibitem[{Bai et~al.(2018)Bai, Kolter, and Koltun}]{bai2018empirical}
Shaojie Bai, J~Zico Kolter, and Vladlen Koltun. 2018.
\newblock An empirical evaluation of generic convolutional and recurrent networks for sequence modeling.
\newblock \emph{arXiv preprint arXiv:1803.01271}.

\bibitem[{Baltru{\v{s}}aitis et~al.(2018)Baltru{\v{s}}aitis, Ahuja, and Morency}]{baltruvsaitis2018multimodal}
Tadas Baltru{\v{s}}aitis, Chaitanya Ahuja, and Louis-Philippe Morency. 2018.
\newblock Multimodal machine learning: A survey and taxonomy.
\newblock \emph{IEEE transactions on pattern analysis and machine intelligence}, 41(2):423--443.

\bibitem[{Bedda and Hammami(2010)}]{misc_spoken_arabic_digit_195}
Mouldi Bedda and Nacereddine Hammami. 2010.
\newblock {Spoken Arabic Digit}.
\newblock UCI Machine Learning Repository.
\newblock {DOI}: https://doi.org/10.24432/C52C9Q.

\bibitem[{Berndt and Clifford(1994)}]{berndt1994using}
Donald~J Berndt and James Clifford. 1994.
\newblock Using dynamic time warping to find patterns in time series.
\newblock In \emph{Proceedings of the 3rd international conference on knowledge discovery and data mining}, pages 359--370.

\bibitem[{Birbaumer et~al.(1999)Birbaumer, Ghanayim, Hinterberger, Iversen, Kotchoubey, K{\"u}bler, Perelmouter, Taub, and Flor}]{birbaumer1999spelling}
Niels Birbaumer, Nimr Ghanayim, Thilo Hinterberger, Iver Iversen, Boris Kotchoubey, Andrea K{\"u}bler, Juri Perelmouter, Edward Taub, and Herta Flor. 1999.
\newblock A spelling device for the paralysed.
\newblock \emph{Nature}, 398(6725):297--298.

\bibitem[{Bommasani et~al.(2021)Bommasani, Hudson, Adeli, Altman, Arora, von Arx, Bernstein, Bohg, Bosselut, Brunskill et~al.}]{bommasani2021opportunities}
Rishi Bommasani, Drew~A Hudson, Ehsan Adeli, Russ Altman, Simran Arora, Sydney von Arx, Michael~S Bernstein, Jeannette Bohg, Antoine Bosselut, Emma Brunskill, et~al. 2021.
\newblock On the opportunities and risks of foundation models.
\newblock \emph{arXiv preprint arXiv:2108.07258}.

\bibitem[{Cao et~al.(2023)Cao, Jia, Arik, Pfister, Zheng, Ye, and Liu}]{cao2023tempo}
Defu Cao, Furong Jia, Sercan~O Arik, Tomas Pfister, Yixiang Zheng, Wen Ye, and Yan Liu. 2023.
\newblock Tempo: Prompt-based generative pre-trained transformer for time series forecasting.
\newblock \emph{arXiv preprint arXiv:2310.04948}.

\bibitem[{Chen and Guestrin(2016)}]{chen2016xgboost}
Tianqi Chen and Carlos Guestrin. 2016.
\newblock Xgboost: A scalable tree boosting system.
\newblock In \emph{Proceedings of the 22nd acm sigkdd international conference on knowledge discovery and data mining}, pages 785--794.

\bibitem[{Chu et~al.(2024)Chu, Chen, Chen, Yu, Wang, Liu, and Qin}]{chu-etal-2024-timebench}
Zheng Chu, Jingchang Chen, Qianglong Chen, Weijiang Yu, Haotian Wang, Ming Liu, and Bing Qin. 2024.
\newblock \href {https://doi.org/10.18653/v1/2024.acl-long.66} {{T}ime{B}ench: A comprehensive evaluation of temporal reasoning abilities in large language models}.
\newblock In \emph{Proceedings of the 62nd Annual Meeting of the Association for Computational Linguistics (ACL) (Volume 1: Long Papers)}, pages 1204--1228, Bangkok, Thailand. Association for Computational Linguistics.

\bibitem[{Cuturi(2011)}]{cuturi2011fast}
Marco Cuturi. 2011.
\newblock Fast global alignment kernels.
\newblock In \emph{Proceedings of the 28th international conference on machine learning (ICML-11)}, pages 929--936.

\bibitem[{Dao et~al.(2022)Dao, Fu, Ermon, Rudra, and R{\'e}}]{dao2022flashattention}
Tri Dao, Dan Fu, Stefano Ermon, Atri Rudra, and Christopher R{\'e}. 2022.
\newblock Flashattention: Fast and memory-efficient exact attention with io-awareness.
\newblock \emph{Advances in Neural Information Processing Systems}, 35:16344--16359.

\bibitem[{Dempster et~al.(2020)Dempster, Petitjean, and Webb}]{dempster2020rocket}
Angus Dempster, Fran{\c{c}}ois Petitjean, and Geoffrey~I Webb. 2020.
\newblock Rocket: exceptionally fast and accurate time series classification using random convolutional kernels.
\newblock \emph{Data Mining and Knowledge Discovery}, 34(5):1454--1495.

\bibitem[{Devlin et~al.(2019)Devlin, Chang, Lee, and Toutanova}]{devlin-etal-2019-bert}
Jacob Devlin, Ming-Wei Chang, Kenton Lee, and Kristina Toutanova. 2019.
\newblock \href {https://doi.org/10.18653/v1/N19-1423} {{BERT}: Pre-training of deep bidirectional transformers for language understanding}.
\newblock In \emph{Proceedings of the 2019 Conference of the North {A}merican Chapter of the Association for Computational Linguistics (NAACL): Human Language Technologies, Volume 1 (Long and Short Papers)}, pages 4171--4186, Minneapolis, Minnesota. Association for Computational Linguistics.

\bibitem[{Deznabi et~al.(2021)Deznabi, Iyyer, and Fiterau}]{deznabi2021predicting}
Iman Deznabi, Mohit Iyyer, and Madalina Fiterau. 2021.
\newblock Predicting in-hospital mortality by combining clinical notes with time-series data.
\newblock In \emph{Findings of the association for computational linguistics: ACL-IJCNLP 2021}, pages 4026--4031.

\bibitem[{Drinkall et~al.(2024)Drinkall, Rahimikia, Pierrehumbert, and Zohren}]{drinkall-etal-2024-time}
Felix Drinkall, Eghbal Rahimikia, Janet Pierrehumbert, and Stefan Zohren. 2024.
\newblock \href {https://doi.org/10.18653/v1/2024.findings-naacl.208} {Time machine {GPT}}.
\newblock In \emph{Findings of the Association for Computational Linguistics: NAACL 2024}, pages 3281--3292, Mexico City, Mexico. Association for Computational Linguistics.

\bibitem[{Eldele et~al.(2021)Eldele, Ragab, Chen, Wu, Kwoh, Li, and Guan}]{eldele2021time}
Emadeldeen Eldele, Mohamed Ragab, Zhenghua Chen, Min Wu, Chee~Keong Kwoh, Xiaoli Li, and Cuntai Guan. 2021.
\newblock Time-series representation learning via temporal and contextual contrasting.
\newblock In \emph{Proceedings of the Thirtieth International Joint Conference on Artificial Intelligence}. International Joint Conferences on Artificial Intelligence Organization.

\bibitem[{Fons et~al.(2024)Fons, Kaur, Palande, Zeng, Balch, Veloso, and Vyetrenko}]{fons-etal-2024-evaluating}
Elizabeth Fons, Rachneet Kaur, Soham Palande, Zhen Zeng, Tucker Balch, Manuela Veloso, and Svitlana Vyetrenko. 2024.
\newblock \href {https://doi.org/10.18653/v1/2024.emnlp-main.1204} {Evaluating large language models on time series feature understanding: A comprehensive taxonomy and benchmark}.
\newblock In \emph{Proceedings of the 2024 Conference on Empirical Methods in Natural Language Processing (EMNLP)}, pages 21598--21634, Miami, Florida, USA. Association for Computational Linguistics.

\bibitem[{Franceschi et~al.(2019)Franceschi, Dieuleveut, and Jaggi}]{franceschi2019unsupervised}
Jean-Yves Franceschi, Aymeric Dieuleveut, and Martin Jaggi. 2019.
\newblock Unsupervised scalable representation learning for multivariate time series.
\newblock \emph{Advances in neural information processing systems}, 32.

\bibitem[{Geiping and Goldstein(2023)}]{geiping2023cramming}
Jonas Geiping and Tom Goldstein. 2023.
\newblock Cramming: Training a language model on a single gpu in one day.
\newblock In \emph{International Conference on Machine Learning}, pages 11117--11143. PMLR.

\bibitem[{Gomez et~al.(2017)Gomez, Ren, Urtasun, and Grosse}]{gomez2017reversible}
Aidan~N Gomez, Mengye Ren, Raquel Urtasun, and Roger~B Grosse. 2017.
\newblock The reversible residual network: Backpropagation without storing activations.
\newblock \emph{Advances in neural information processing systems}, 30.

\bibitem[{Goswami et~al.(2024)Goswami, Szafer, Choudhry, Cai, Li, and Dubrawski}]{goswamimoment}
Mononito Goswami, Konrad Szafer, Arjun Choudhry, Yifu Cai, Shuo Li, and Artur Dubrawski. 2024.
\newblock Moment: A family of open time-series foundation models.
\newblock In \emph{Forty-first International Conference on Machine Learning}.

\bibitem[{Gruver et~al.(2024)Gruver, Finzi, Qiu, and Wilson}]{gruver2024large}
Nate Gruver, Marc Finzi, Shikai Qiu, and Andrew~G Wilson. 2024.
\newblock Large language models are zero-shot time series forecasters.
\newblock \emph{Advances in Neural Information Processing Systems}, 36.

\bibitem[{Gu et~al.(2020)Gu, Dao, Ermon, Rudra, and R{\'e}}]{gu2020hippo}
Albert Gu, Tri Dao, Stefano Ermon, Atri Rudra, and Christopher R{\'e}. 2020.
\newblock Hippo: Recurrent memory with optimal polynomial projections.
\newblock \emph{Advances in neural information processing systems}, 33:1474--1487.

\bibitem[{Gu et~al.(2021)Gu, Goel, and R{\'e}}]{gu2021efficiently}
Albert Gu, Karan Goel, and Christopher R{\'e}. 2021.
\newblock Efficiently modeling long sequences with structured state spaces.
\newblock \emph{arXiv preprint arXiv:2111.00396}.

\bibitem[{Guo et~al.(2019)Guo, Wang, and Wang}]{guo2019deep}
Wenzhong Guo, Jianwen Wang, and Shiping Wang. 2019.
\newblock Deep multimodal representation learning: A survey.
\newblock \emph{Ieee Access}, 7:63373--63394.

\bibitem[{He et~al.(2016)He, Zhang, Ren, and Sun}]{he2016deep}
Kaiming He, Xiangyu Zhang, Shaoqing Ren, and Jian Sun. 2016.
\newblock Deep residual learning for image recognition.
\newblock In \emph{Proceedings of the IEEE conference on computer vision and pattern recognition}, pages 770--778.

\bibitem[{Hochreiter and Schmidhuber(1997)}]{hochreiter1997long}
Sepp Hochreiter and J{\"u}rgen Schmidhuber. 1997.
\newblock Long short-term memory.
\newblock \emph{Neural computation}, 9(8):1735--1780.

\bibitem[{Hu et~al.(2021)Hu, Shen, Wallis, Allen-Zhu, Li, Wang, Wang, and Chen}]{hu2021lora}
Edward~J Hu, Yelong Shen, Phillip Wallis, Zeyuan Allen-Zhu, Yuanzhi Li, Shean Wang, Lu~Wang, and Weizhu Chen. 2021.
\newblock Lora: Low-rank adaptation of large language models.
\newblock \emph{arXiv preprint arXiv:2106.09685}.

\bibitem[{Ismail~Fawaz et~al.(2019)Ismail~Fawaz, Forestier, Weber, Idoumghar, and Muller}]{ismail2019deep}
Hassan Ismail~Fawaz, Germain Forestier, Jonathan Weber, Lhassane Idoumghar, and Pierre-Alain Muller. 2019.
\newblock Deep learning for time series classification: a review.
\newblock \emph{Data mining and knowledge discovery}, 33(4):917--963.

\bibitem[{Jabri et~al.(2016)Jabri, Joulin, and Van Der~Maaten}]{jabri2016revisiting}
Allan Jabri, Armand Joulin, and Laurens Van Der~Maaten. 2016.
\newblock Revisiting visual question answering baselines.
\newblock In \emph{European conference on computer vision}, pages 727--739. Springer.

\bibitem[{Jeong et~al.(2011)Jeong, Jeong, and Omitaomu}]{jeong2011weighted}
Young-Seon Jeong, Myong~K Jeong, and Olufemi~A Omitaomu. 2011.
\newblock Weighted dynamic time warping for time series classification.
\newblock \emph{Pattern recognition}, 44(9):2231--2240.

\bibitem[{Jhamtani and Berg-Kirkpatrick(2021)}]{jhamtani-berg-kirkpatrick-2021-truth}
Harsh Jhamtani and Taylor Berg-Kirkpatrick. 2021.
\newblock \href {https://doi.org/10.18653/v1/2021.emnlp-main.55} {Truth-conditional captions for time series data}.
\newblock In \emph{Proceedings of the 2021 Conference on Empirical Methods in Natural Language Processing (EMNLP)}, pages 719--733, Online and Punta Cana, Dominican Republic. Association for Computational Linguistics.

\bibitem[{Jiang et~al.(2023)Jiang, Sablayrolles, Mensch, Bamford, Chaplot, Casas, Bressand, Lengyel, Lample, Saulnier et~al.}]{jiang2023mistral}
Albert~Q Jiang, Alexandre Sablayrolles, Arthur Mensch, Chris Bamford, Devendra~Singh Chaplot, Diego de~las Casas, Florian Bressand, Gianna Lengyel, Guillaume Lample, Lucile Saulnier, et~al. 2023.
\newblock Mistral 7b.
\newblock \emph{arXiv preprint arXiv:2310.06825}.

\bibitem[{Jiang et~al.(2024)Jiang, Pan, Zhang, Garg, Schneider, Nevmyvaka, and Song}]{Jiang2024EmpoweringTS}
Yushan Jiang, Zijie Pan, Xikun Zhang, Sahil Garg, Anderson Schneider, Yuriy Nevmyvaka, and Dongjin Song. 2024.
\newblock \href {https://api.semanticscholar.org/CorpusID:267412144} {Empowering time series analysis with large language models: A survey}.
\newblock \emph{ArXiv}, abs/2402.03182.

\bibitem[{Jin et~al.(2023)Jin, Wang, Ma, Chu, Zhang, Shi, Chen, Liang, Li, Pan et~al.}]{jin2023time}
Ming Jin, Shiyu Wang, Lintao Ma, Zhixuan Chu, James~Y Zhang, Xiaoming Shi, Pin-Yu Chen, Yuxuan Liang, Yuan-Fang Li, Shirui Pan, et~al. 2023.
\newblock Time-llm: Time series forecasting by reprogramming large language models.
\newblock \emph{arXiv preprint arXiv:2310.01728}.

\bibitem[{Kampouraki et~al.(2008)Kampouraki, Manis, and Nikou}]{kampouraki2008heartbeat}
Argyro Kampouraki, George Manis, and Christophoros Nikou. 2008.
\newblock Heartbeat time series classification with support vector machines.
\newblock \emph{IEEE transactions on information technology in biomedicine}, 13(4):512--518.

\bibitem[{Kawarada et~al.(2024)Kawarada, Ishigaki, Topi{\'c}, and Takamura}]{kawarada-etal-2024-demonstration}
Masayuki Kawarada, Tatsuya Ishigaki, Goran Topi{\'c}, and Hiroya Takamura. 2024.
\newblock \href {https://doi.org/10.18653/v1/2024.findings-emnlp.435} {Demonstration selection strategies for numerical time series data-to-text}.
\newblock In \emph{Findings of the Association for Computational Linguistics: EMNLP 2024}, pages 7378--7392, Miami, Florida, USA. Association for Computational Linguistics.

\bibitem[{Ke et~al.(2017)Ke, Meng, Finley, Wang, Chen, Ma, Ye, and Liu}]{ke2017lightgbm}
Guolin Ke, Qi~Meng, Thomas Finley, Taifeng Wang, Wei Chen, Weidong Ma, Qiwei Ye, and Tie-Yan Liu. 2017.
\newblock Lightgbm: A highly efficient gradient boosting decision tree.
\newblock \emph{Advances in neural information processing systems}, 30.

\bibitem[{Khadanga et~al.(2019)Khadanga, Aggarwal, Joty, and Srivastava}]{khadanga-etal-2019-using}
Swaraj Khadanga, Karan Aggarwal, Shafiq Joty, and Jaideep Srivastava. 2019.
\newblock \href {https://doi.org/10.18653/v1/D19-1678} {Using clinical notes with time series data for {ICU} management}.
\newblock In \emph{Proceedings of the 2019 Conference on Empirical Methods in Natural Language Processing and the 9th International Joint Conference on Natural Language Processing (EMNLP-IJCNLP)}, pages 6432--6437, Hong Kong, China. Association for Computational Linguistics.

\bibitem[{Kim et~al.(2021)Kim, Kim, Tae, Park, Choi, and Choo}]{kim2021reversible}
Taesung Kim, Jinhee Kim, Yunwon Tae, Cheonbok Park, Jang-Ho Choi, and Jaegul Choo. 2021.
\newblock Reversible instance normalization for accurate time-series forecasting against distribution shift.
\newblock In \emph{International Conference on Learning Representations}.

\bibitem[{Kitaev et~al.(2020)Kitaev, Kaiser, and Levskaya}]{kitaev2020reformer}
Nikita Kitaev, {\L}ukasz Kaiser, and Anselm Levskaya. 2020.
\newblock Reformer: The efficient transformer.
\newblock \emph{arXiv preprint arXiv:2001.04451}.

\bibitem[{Koval et~al.(2024)Koval, Andrews, and Yan}]{koval-etal-2024-financial}
Ross Koval, Nicholas Andrews, and Xifeng Yan. 2024.
\newblock \href {https://doi.org/10.18653/v1/2024.findings-emnlp.486} {Financial forecasting from textual and tabular time series}.
\newblock In \emph{Findings of the Association for Computational Linguistics: EMNLP 2024}, pages 8289--8300, Miami, Florida, USA. Association for Computational Linguistics.

\bibitem[{Kudo et~al.(1999)Kudo, Toyama, and Shimbo}]{misc_japanese_vowels_128}
Mineichi Kudo, Jun Toyama, and Masaru Shimbo. 1999.
\newblock {Japanese Vowels}.
\newblock UCI Machine Learning Repository.
\newblock {DOI}: https://doi.org/10.24432/C5NS47.

\bibitem[{Kusupati et~al.(2022)Kusupati, Bhatt, Rege, Wallingford, Sinha, Ramanujan, Howard-Snyder, Chen, Kakade, Jain et~al.}]{kusupati2022matryoshka}
Aditya Kusupati, Gantavya Bhatt, Aniket Rege, Matthew Wallingford, Aditya Sinha, Vivek Ramanujan, William Howard-Snyder, Kaifeng Chen, Sham Kakade, Prateek Jain, et~al. 2022.
\newblock Matryoshka representation learning.
\newblock \emph{Advances in Neural Information Processing Systems}, 35:30233--30249.

\bibitem[{Lai et~al.(2018)Lai, Chang, Yang, and Liu}]{lai2018modeling}
Guokun Lai, Wei-Cheng Chang, Yiming Yang, and Hanxiao Liu. 2018.
\newblock Modeling long-and short-term temporal patterns with deep neural networks.
\newblock In \emph{The 41st international ACM SIGIR conference on research \& development in information retrieval}, pages 95--104.

\bibitem[{Large et~al.(2018)Large, Kemsley, Wellner, Goodall, and Bagnall}]{large2018detecting}
James Large, E~Kate Kemsley, Nikolaus Wellner, Ian Goodall, and Anthony Bagnall. 2018.
\newblock Detecting forged alcohol non-invasively through vibrational spectroscopy and machine learning.
\newblock In \emph{Pacific-Asia Conference on Knowledge Discovery and Data Mining}, pages 298--309. Springer.

\bibitem[{Levy et~al.(2024)Levy, Jacoby, and Goldberg}]{levy2024same}
Mosh Levy, Alon Jacoby, and Yoav Goldberg. 2024.
\newblock Same task, more tokens: the impact of input length on the reasoning performance of large language models.
\newblock \emph{arXiv preprint arXiv:2402.14848}.

\bibitem[{Li et~al.(2024{\natexlab{a}})Li, Gao, Li, Li, and Liao}]{li-etal-2024-econagent}
Nian Li, Chen Gao, Mingyu Li, Yong Li, and Qingmin Liao. 2024{\natexlab{a}}.
\newblock \href {https://doi.org/10.18653/v1/2024.acl-long.829} {{E}con{A}gent: Large language model-empowered agents for simulating macroeconomic activities}.
\newblock In \emph{Proceedings of the 62nd Annual Meeting of the Association for Computational Linguistics (ACL) (Volume 1: Long Papers)}, pages 15523--15536, Bangkok, Thailand. Association for Computational Linguistics.

\bibitem[{Li et~al.(2023)Li, Zhang, Zhang, Long, Xie, and Zhang}]{li2023towards}
Zehan Li, Xin Zhang, Yanzhao Zhang, Dingkun Long, Pengjun Xie, and Meishan Zhang. 2023.
\newblock Towards general text embeddings with multi-stage contrastive learning.
\newblock \emph{arXiv preprint arXiv:2308.03281}.

\bibitem[{Li et~al.(2024{\natexlab{b}})Li, Li, and Yan}]{li2024time}
Zekun Li, Shiyang Li, and Xifeng Yan. 2024{\natexlab{b}}.
\newblock Time series as images: Vision transformer for irregularly sampled time series.
\newblock \emph{Advances in Neural Information Processing Systems}, 36.

\bibitem[{Lipton et~al.(2016)Lipton, Kale, and Wetzel}]{lipton2016directly}
Zachary~C Lipton, David Kale, and Randall Wetzel. 2016.
\newblock Directly modeling missing data in sequences with rnns: Improved classification of clinical time series.
\newblock In \emph{Machine learning for healthcare conference}, pages 253--270. PMLR.

\bibitem[{Liu et~al.(2016)Liu, Springer, Li, Moody, Juan, Chorro, Castells, Roig, Silva, Johnson et~al.}]{liu2016open}
Chengyu Liu, David Springer, Qiao Li, Benjamin Moody, Ricardo~Abad Juan, Francisco~J Chorro, Francisco Castells, Jos{\'e}~Millet Roig, Ikaro Silva, Alistair~EW Johnson, et~al. 2016.
\newblock An open access database for the evaluation of heart sound algorithms.
\newblock \emph{Physiological measurement}, 37(12):2181.

\bibitem[{Liu et~al.(2024)Liu, Zhao, Wang, Kamarthi, and Prakash}]{liu-etal-2024-lstprompt}
Haoxin Liu, Zhiyuan Zhao, Jindong Wang, Harshavardhan Kamarthi, and B.~Aditya Prakash. 2024.
\newblock \href {https://doi.org/10.18653/v1/2024.findings-acl.466} {{LSTP}rompt: Large language models as zero-shot time series forecasters by long-short-term prompting}.
\newblock In \emph{Findings of the Association for Computational Linguistics: ACL 2024}, pages 7832--7840, Bangkok, Thailand. Association for Computational Linguistics.

\bibitem[{Liu et~al.(2009)Liu, Zhong, Wickramasuriya, and Vasudevan}]{liu2009uwave}
Jiayang Liu, Lin Zhong, Jehan Wickramasuriya, and Venu Vasudevan. 2009.
\newblock uwave: Accelerometer-based personalized gesture recognition and its applications.
\newblock \emph{Pervasive and Mobile Computing}, 5(6):657--675.

\bibitem[{Liu et~al.(2019{\natexlab{a}})Liu, Jiang, He, Chen, Liu, Gao, and Han}]{liu2019variance}
Liyuan Liu, Haoming Jiang, Pengcheng He, Weizhu Chen, Xiaodong Liu, Jianfeng Gao, and Jiawei Han. 2019{\natexlab{a}}.
\newblock On the variance of the adaptive learning rate and beyond.
\newblock \emph{arXiv preprint arXiv:1908.03265}.

\bibitem[{Liu et~al.(2021)Liu, Yu, Liao, Li, Lin, Liu, and Dustdar}]{liu2021pyraformer}
Shizhan Liu, Hang Yu, Cong Liao, Jianguo Li, Weiyao Lin, Alex~X Liu, and Schahram Dustdar. 2021.
\newblock Pyraformer: Low-complexity pyramidal attention for long-range time series modeling and forecasting.
\newblock In \emph{International conference on learning representations}.

\bibitem[{Liu and Low(2023)}]{liu2023goat}
Tiedong Liu and Bryan Kian~Hsiang Low. 2023.
\newblock Goat: Fine-tuned llama outperforms gpt-4 on arithmetic tasks.
\newblock \emph{arXiv preprint arXiv:2305.14201}.

\bibitem[{Liu et~al.(2019{\natexlab{b}})Liu, Ott, Goyal, Du, Joshi, Chen, Levy, Lewis, Zettlemoyer, and Stoyanov}]{liu2019roberta}
Yinhan Liu, Myle Ott, Naman Goyal, Jingfei Du, Mandar Joshi, Danqi Chen, Omer Levy, Mike Lewis, Luke Zettlemoyer, and Veselin Stoyanov. 2019{\natexlab{b}}.
\newblock Roberta: A robustly optimized bert pretraining approach.
\newblock \emph{arXiv preprint arXiv:1907.11692}.

\bibitem[{Liu et~al.(2022)Liu, Wu, Wang, and Long}]{liu2022non}
Yong Liu, Haixu Wu, Jianmin Wang, and Mingsheng Long. 2022.
\newblock Non-stationary transformers: Exploring the stationarity in time series forecasting.
\newblock \emph{Advances in Neural Information Processing Systems}, 35:9881--9893.

\bibitem[{Ma et~al.(2024)Ma, Wang, Yang, Wei, and Lin}]{ma2024fine}
Xueguang Ma, Liang Wang, Nan Yang, Furu Wei, and Jimmy Lin. 2024.
\newblock Fine-tuning llama for multi-stage text retrieval.
\newblock In \emph{Proceedings of the 47th International ACM SIGIR Conference on Research and Development in Information Retrieval}, pages 2421--2425.

\bibitem[{Manzoor et~al.(2023)Manzoor, Albarri, Xian, Meng, Nakov, and Liang}]{manzoor2023multimodality}
Muhammad~Arslan Manzoor, Sarah Albarri, Ziting Xian, Zaiqiao Meng, Preslav Nakov, and Shangsong Liang. 2023.
\newblock Multimodality representation learning: A survey on evolution, pretraining and its applications.
\newblock \emph{ACM Transactions on Multimedia Computing, Communications and Applications}, 20(3):1--34.

\bibitem[{Merrill et~al.(2024)Merrill, Tan, Gupta, Hartvigsen, and Althoff}]{merrill-etal-2024-language}
Mike~A Merrill, Mingtian Tan, Vinayak Gupta, Thomas Hartvigsen, and Tim Althoff. 2024.
\newblock \href {https://doi.org/10.18653/v1/2024.findings-emnlp.201} {Language models still struggle to zero-shot reason about time series}.
\newblock In \emph{Findings of the Association for Computational Linguistics: EMNLP 2024}, pages 3512--3533, Miami, Florida, USA. Association for Computational Linguistics.

\bibitem[{Mikolov et~al.(2013)Mikolov, Sutskever, Chen, Corrado, and Dean}]{mikolov2013distributed}
Tomas Mikolov, Ilya Sutskever, Kai Chen, Greg~S Corrado, and Jeff Dean. 2013.
\newblock Distributed representations of words and phrases and their compositionality.
\newblock \emph{Advances in neural information processing systems}, 26.

\bibitem[{MTEB(2024)}]{MTEBleaderboard}
MTEB. 2024.
\newblock {M}assive {T}ext {E}mbedding {B}enchmark ({MTEB}) {L}eaderboard.
\newblock \url{https://huggingface.co/spaces/mteb/leaderboard}.
\newblock Accessed: 2024-05-13.

\bibitem[{Muennighoff et~al.(2022)Muennighoff, Tazi, Magne, and Reimers}]{muennighoff2022mteb}
Niklas Muennighoff, Nouamane Tazi, Lo{\"\i}c Magne, and Nils Reimers. 2022.
\newblock \href {https://doi.org/10.48550/ARXIV.2210.07316} {Mteb: Massive text embedding benchmark}.
\newblock \emph{arXiv preprint arXiv:2210.07316}.

\bibitem[{Murakami et~al.(2017)Murakami, Watanabe, Miyazawa, Goshima, Yanase, Takamura, and Miyao}]{murakami2017learning}
Soichiro Murakami, Akihiko Watanabe, Akira Miyazawa, Keiichi Goshima, Toshihiko Yanase, Hiroya Takamura, and Yusuke Miyao. 2017.
\newblock Learning to generate market comments from stock prices.
\newblock In \emph{Proceedings of the 55th Annual Meeting of the Association for Computational Linguistics (ACL) (Volume 1: Long Papers)}, pages 1374--1384.

\bibitem[{Nie et~al.(2023)Nie, Nguyen, Sinthong, and Kalagnanam}]{nietime}
Yuqi Nie, Nam~H Nguyen, Phanwadee Sinthong, and Jayant Kalagnanam. 2023.
\newblock A time series is worth 64 words: Long-term forecasting with transformers.
\newblock In \emph{The Eleventh International Conference on Learning Representations}.

\bibitem[{Nussbaum et~al.(2024)Nussbaum, Morris, Duderstadt, and Mulyar}]{nussbaum2024nomic}
Zach Nussbaum, John~X. Morris, Brandon Duderstadt, and Andriy Mulyar. 2024.
\newblock \href {https://arxiv.org/abs/2402.01613} {Nomic embed: Training a reproducible long context text embedder}.
\newblock \emph{Preprint}, arXiv:2402.01613.

\bibitem[{OpenAI(2024)}]{openai_large_embeddings}
OpenAI. 2024.
\newblock {O}pen{AI} {E}mbedding {M}odels.
\newblock \url{https://platform.openai.com/docs/guides/embeddings}, \url{https://openai.com/index/new-embedding-models-and-api-updates/}.
\newblock Accessed: 2024-05-13.

\bibitem[{Pan(2017)}]{pan2017fast}
Victor Pan. 2017.
\newblock Fast approximate computations with cauchy matrices and polynomials.
\newblock \emph{Mathematics of Computation}, 86(308):2799--2826.

\bibitem[{Pan(2012)}]{pan2012structured}
Victor~Y Pan. 2012.
\newblock \emph{Structured matrices and polynomials: unified superfast algorithms}.
\newblock Springer Science \& Business Media.

\bibitem[{Passalis et~al.(2017)Passalis, Tsantekidis, Tefas, Kanniainen, Gabbouj, and Iosifidis}]{passalis2017time}
Nikolaos Passalis, Avraam Tsantekidis, Anastasios Tefas, Juho Kanniainen, Moncef Gabbouj, and Alexandros Iosifidis. 2017.
\newblock Time-series classification using neural bag-of-features.
\newblock In \emph{2017 25th European Signal Processing Conference (EUSIPCO)}, pages 301--305. IEEE.

\bibitem[{Pennington et~al.(2014)Pennington, Socher, and Manning}]{pennington2014glove}
Jeffrey Pennington, Richard Socher, and Christopher~D Manning. 2014.
\newblock Glove: Global vectors for word representation.
\newblock In \emph{Proceedings of the 2014 conference on empirical methods in natural language processing (EMNLP)}, pages 1532--1543.

\bibitem[{Poria et~al.(2018)Poria, Majumder, Hazarika, Cambria, Gelbukh, and Hussain}]{poria2018multimodal}
Soujanya Poria, Navonil Majumder, Devamanyu Hazarika, Erik Cambria, Alexander Gelbukh, and Amir Hussain. 2018.
\newblock Multimodal sentiment analysis: Addressing key issues and setting up the baselines.
\newblock \emph{IEEE Intelligent Systems}, 33(6):17--25.

\bibitem[{Portes et~al.(2024)Portes, Trott, Havens, King, Venigalla, Nadeem, Sardana, Khudia, and Frankle}]{portes2024mosaicbert}
Jacob Portes, Alexander Trott, Sam Havens, Daniel King, Abhinav Venigalla, Moin Nadeem, Nikhil Sardana, Daya Khudia, and Jonathan Frankle. 2024.
\newblock Mosaicbert: a bidirectional encoder optimized for fast pretraining.
\newblock \emph{Advances in Neural Information Processing Systems}, 36.

\bibitem[{Press et~al.(2020)Press, Smith, and Lewis}]{press2020shortformer}
Ofir Press, Noah~A Smith, and Mike Lewis. 2020.
\newblock Shortformer: Better language modeling using shorter inputs.
\newblock \emph{arXiv preprint arXiv:2012.15832}.

\bibitem[{Radford et~al.(2019)Radford, Wu, Child, Luan, Amodei, and Sutskever}]{radford2019language}
Alec Radford, Jeff Wu, Rewon Child, David Luan, Dario Amodei, and Ilya Sutskever. 2019.
\newblock Language models are unsupervised multitask learners.

\bibitem[{Rik~Henson(2023)}]{face_detection_dataset}
UEA Rik~Henson. 2023.
\newblock {D}ec{M}eg2014 - {D}ecoding the {H}uman {B}rain.
\newblock \url{https://www.kaggle.com/c/decoding-the-human-brain/data}.
\newblock Accessed: 2024-04-15.

\bibitem[{Salimans and Kingma(2016)}]{salimans2016weight}
Tim Salimans and Durk~P Kingma. 2016.
\newblock Weight normalization: A simple reparameterization to accelerate training of deep neural networks.
\newblock \emph{Advances in neural information processing systems}, 29.

\bibitem[{Shazeer(2020)}]{shazeer2020glu}
Noam Shazeer. 2020.
\newblock Glu variants improve transformer.
\newblock \emph{arXiv preprint arXiv:2002.05202}.

\bibitem[{Shimodaira et~al.(2001)Shimodaira, Noma, Nakai, and Sagayama}]{shimodaira2001dynamic}
Hiroshi Shimodaira, Ken-ichi Noma, Mitsuru Nakai, and Shigeki Sagayama. 2001.
\newblock Dynamic time-alignment kernel in support vector machine.
\newblock \emph{Advances in neural information processing systems}, 14.

\bibitem[{Shoeybi et~al.(2019)Shoeybi, Patwary, Puri, LeGresley, Casper, and Catanzaro}]{shoeybi2019megatron}
Mohammad Shoeybi, Mostofa Patwary, Raul Puri, Patrick LeGresley, Jared Casper, and Bryan Catanzaro. 2019.
\newblock Megatron-lm: Training multi-billion parameter language models using model parallelism.
\newblock \emph{arXiv preprint arXiv:1909.08053}.

\bibitem[{Shokoohi-Yekta et~al.(2017)Shokoohi-Yekta, Hu, Jin, Wang, and Keogh}]{shokoohi2017generalizing}
Mohammad Shokoohi-Yekta, Bing Hu, Hongxia Jin, Jun Wang, and Eamonn Keogh. 2017.
\newblock Generalizing dtw to the multi-dimensional case requires an adaptive approach.
\newblock \emph{Data mining and knowledge discovery}, 31:1--31.

\bibitem[{Su et~al.(2024)Su, Ahmed, Lu, Pan, Bo, and Liu}]{su2024roformer}
Jianlin Su, Murtadha Ahmed, Yu~Lu, Shengfeng Pan, Wen Bo, and Yunfeng Liu. 2024.
\newblock Roformer: Enhanced transformer with rotary position embedding.
\newblock \emph{Neurocomputing}, 568:127063.

\bibitem[{Sun et~al.(2023)Sun, Li, Li, and Hong}]{sun2023test}
Chenxi Sun, Yaliang Li, Hongyan Li, and Shenda Hong. 2023.
\newblock Test: Text prototype aligned embedding to activate llm's ability for time series.
\newblock \emph{arXiv preprint arXiv:2308.08241}.

\bibitem[{Szegedy et~al.(2015)Szegedy, Liu, Jia, Sermanet, Reed, Anguelov, Erhan, Vanhoucke, and Rabinovich}]{szegedy2015going}
Christian Szegedy, Wei Liu, Yangqing Jia, Pierre Sermanet, Scott Reed, Dragomir Anguelov, Dumitru Erhan, Vincent Vanhoucke, and Andrew Rabinovich. 2015.
\newblock Going deeper with convolutions.
\newblock In \emph{Proceedings of the IEEE conference on computer vision and pattern recognition}, pages 1--9.

\bibitem[{Tonekaboni et~al.(2021)Tonekaboni, Eytan, and Goldenberg}]{tonekaboniunsupervised}
Sana Tonekaboni, Danny Eytan, and Anna Goldenberg. 2021.
\newblock Unsupervised representation learning for time series with temporal neighborhood coding.
\newblock In \emph{International Conference on Learning Representations (ICLR)}.

\bibitem[{Vaswani et~al.(2017)Vaswani, Shazeer, Parmar, Uszkoreit, Jones, Gomez, Kaiser, and Polosukhin}]{vaswani2017attention}
Ashish Vaswani, Noam Shazeer, Niki Parmar, Jakob Uszkoreit, Llion Jones, Aidan~N Gomez, {\L}ukasz Kaiser, and Illia Polosukhin. 2017.
\newblock Attention is all you need.
\newblock \emph{Advances in neural information processing systems}, 30.

\bibitem[{Voelker et~al.(2019)Voelker, Kaji{\'c}, and Eliasmith}]{voelker2019legendre}
Aaron Voelker, Ivana Kaji{\'c}, and Chris Eliasmith. 2019.
\newblock Legendre memory units: Continuous-time representation in recurrent neural networks.
\newblock \emph{Advances in neural information processing systems}, 32.

\bibitem[{Wang et~al.(2024)Wang, Yang, Huang, Yang, Majumder, and Wei}]{wang-etal-2024-improving-text}
Liang Wang, Nan Yang, Xiaolong Huang, Linjun Yang, Rangan Majumder, and Furu Wei. 2024.
\newblock \href {https://doi.org/10.18653/v1/2024.acl-long.642} {Improving text embeddings with large language models}.
\newblock In \emph{Proceedings of the 62nd Annual Meeting of the Association for Computational Linguistics (ACL) (Volume 1: Long Papers)}, pages 11897--11916, Bangkok, Thailand. Association for Computational Linguistics.

\bibitem[{Wang et~al.(2013)Wang, Mueen, Ding, Trajcevski, Scheuermann, and Keogh}]{wang2013experimental}
Xiaoyue Wang, Abdullah Mueen, Hui Ding, Goce Trajcevski, Peter Scheuermann, and Eamonn Keogh. 2013.
\newblock Experimental comparison of representation methods and distance measures for time series data.
\newblock \emph{Data Mining and Knowledge Discovery}, 26:275--309.

\bibitem[{Woo et~al.(2022)Woo, Liu, Sahoo, Kumar, and Hoi}]{woo2022etsformer}
Gerald Woo, Chenghao Liu, Doyen Sahoo, Akshat Kumar, and Steven Hoi. 2022.
\newblock Etsformer: Exponential smoothing transformers for time-series forecasting.
\newblock \emph{arXiv preprint arXiv:2202.01381}.

\bibitem[{Wu et~al.(2022{\natexlab{a}})Wu, Hu, Liu, Zhou, Wang, and Long}]{wu2022timesnet}
Haixu Wu, Tengge Hu, Yong Liu, Hang Zhou, Jianmin Wang, and Mingsheng Long. 2022{\natexlab{a}}.
\newblock Timesnet: Temporal 2d-variation modeling for general time series analysis.
\newblock In \emph{The eleventh international conference on learning representations}.

\bibitem[{Wu et~al.(2022{\natexlab{b}})Wu, Wu, Xu, Wang, and Long}]{wu2022flowformer}
Haixu Wu, Jialong Wu, Jiehui Xu, Jianmin Wang, and Mingsheng Long. 2022{\natexlab{b}}.
\newblock Flowformer: Linearizing transformers with conservation flows.
\newblock \emph{arXiv preprint arXiv:2202.06258}.

\bibitem[{Wu et~al.(2021)Wu, Xu, Wang, and Long}]{wu2021autoformer}
Haixu Wu, Jiehui Xu, Jianmin Wang, and Mingsheng Long. 2021.
\newblock Autoformer: Decomposition transformers with auto-correlation for long-term series forecasting.
\newblock \emph{Advances in neural information processing systems}, 34:22419--22430.

\bibitem[{Yang et~al.(2021)Yang, Tsai, and Chen}]{yang2021voice2series}
Chao-Han~Huck Yang, Yun-Yun Tsai, and Pin-Yu Chen. 2021.
\newblock Voice2series: Reprogramming acoustic models for time series classification.
\newblock In \emph{International conference on machine learning}, pages 11808--11819. PMLR.

\bibitem[{Yang et~al.(2015)Yang, Nguyen, San, Li, and Krishnaswamy}]{yang2015deep}
Jianbo Yang, Minh~Nhut Nguyen, Phyo~Phyo San, Xiaoli Li, and Shonali Krishnaswamy. 2015.
\newblock Deep convolutional neural networks on multichannel time series for human activity recognition.
\newblock In \emph{Ijcai}, volume~15, pages 3995--4001. Buenos Aires, Argentina.

\bibitem[{Yu et~al.(2023)Yu, Chen, and Lu}]{yu-etal-2023-harnessing}
Xinli Yu, Zheng Chen, and Yanbin Lu. 2023.
\newblock \href {https://doi.org/10.18653/v1/2023.emnlp-industry.69} {Harnessing {LLM}s for temporal data - a study on explainable financial time series forecasting}.
\newblock In \emph{Proceedings of the 2023 Conference on Empirical Methods in Natural Language Processing (EMNLP): Industry Track}, pages 739--753, Singapore. Association for Computational Linguistics.

\bibitem[{Yue et~al.(2022)Yue, Wang, Duan, Yang, Huang, Tong, and Xu}]{yue2022ts2vec}
Zhihan Yue, Yujing Wang, Juanyong Duan, Tianmeng Yang, Congrui Huang, Yunhai Tong, and Bixiong Xu. 2022.
\newblock Ts2vec: Towards universal representation of time series.
\newblock In \emph{Proceedings of the AAAI Conference on Artificial Intelligence}, volume~36, pages 8980--8987.

\bibitem[{Zeng et~al.(2023)Zeng, Chen, Zhang, and Xu}]{zeng2023transformers}
Ailing Zeng, Muxi Chen, Lei Zhang, and Qiang Xu. 2023.
\newblock Are transformers effective for time series forecasting?
\newblock In \emph{Proceedings of the AAAI conference on artificial intelligence}, volume~37, pages 11121--11128.

\bibitem[{Zerveas et~al.(2021)Zerveas, Jayaraman, Patel, Bhamidipaty, and Eickhoff}]{zerveas2021transformer}
George Zerveas, Srideepika Jayaraman, Dhaval Patel, Anuradha Bhamidipaty, and Carsten Eickhoff. 2021.
\newblock A transformer-based framework for multivariate time series representation learning.
\newblock In \emph{Proceedings of the 27th ACM SIGKDD conference on knowledge discovery \& data mining}, pages 2114--2124.

\bibitem[{Zhang et~al.(2022{\natexlab{a}})Zhang, Zhang, Cao, Bian, Yi, Zheng, and Li}]{zhang2022less}
Tianping Zhang, Yizhuo Zhang, Wei Cao, Jiang Bian, Xiaohan Yi, Shun Zheng, and Jian Li. 2022{\natexlab{a}}.
\newblock Less is more: Fast multivariate time series forecasting with light sampling-oriented mlp structures.
\newblock \emph{arXiv preprint arXiv:2207.01186}.

\bibitem[{Zhang et~al.(2022{\natexlab{b}})Zhang, Zhao, Tsiligkaridis, and Zitnik}]{zhang2022self}
Xiang Zhang, Ziyuan Zhao, Theodoros Tsiligkaridis, and Marinka Zitnik. 2022{\natexlab{b}}.
\newblock Self-supervised contrastive pre-training for time series via time-frequency consistency.
\newblock \emph{Advances in Neural Information Processing Systems}, 35:3988--4003.

\bibitem[{Zhang et~al.(2024)Zhang, Chowdhury, Gupta, and Shang}]{Zhang2024LargeLM}
Xiyuan Zhang, Ranak~Roy Chowdhury, Rajesh~K. Gupta, and Jingbo Shang. 2024.
\newblock \href {https://api.semanticscholar.org/CorpusID:267411923} {Large language models for time series: A survey}.
\newblock \emph{ArXiv}, abs/2402.01801.

\bibitem[{Zhao et~al.(2017)Zhao, Lu, Chen, Liu, and Wu}]{zhao2017convolutional}
Bendong Zhao, Huanzhang Lu, Shangfeng Chen, Junliang Liu, and Dongya Wu. 2017.
\newblock Convolutional neural networks for time series classification.
\newblock \emph{Journal of Systems Engineering and Electronics}, 28(1):162--169.

\bibitem[{Zhou et~al.(2021)Zhou, Zhang, Peng, Zhang, Li, Xiong, and Zhang}]{zhou2021informer}
Haoyi Zhou, Shanghang Zhang, Jieqi Peng, Shuai Zhang, Jianxin Li, Hui Xiong, and Wancai Zhang. 2021.
\newblock Informer: Beyond efficient transformer for long sequence time-series forecasting.
\newblock In \emph{Proceedings of the AAAI conference on artificial intelligence}, volume~35, pages 11106--11115.

\bibitem[{Zhou et~al.(2022)Zhou, Ma, Wen, Wang, Sun, and Jin}]{zhou2022fedformer}
Tian Zhou, Ziqing Ma, Qingsong Wen, Xue Wang, Liang Sun, and Rong Jin. 2022.
\newblock Fedformer: Frequency enhanced decomposed transformer for long-term series forecasting.
\newblock In \emph{International conference on machine learning}, pages 27268--27286. PMLR.

\bibitem[{Zhou et~al.(2024)Zhou, Niu, Sun, Jin et~al.}]{zhou2024one}
Tian Zhou, Peisong Niu, Liang Sun, Rong Jin, et~al. 2024.
\newblock One fits all: Power general time series analysis by pretrained lm.
\newblock \emph{Advances in neural information processing systems}, 36.

\end{thebibliography}

\newpage 
\appendix
\section*{Appendix} 
\section*{Table of Contents} 
\setcounter{tocdepth}{1}  
\startcontents[sections] 
\printcontents[sections]{}{1}{}

\section{Related Works}\label{appendix:related_works}
Time series classification has been an active research area for decades. Early approaches investigated distance-based approaches~\cite{abanda2019review} for time series classification. Some built nearest neighbor classifiers based on explicit time series distance measures such as Dynamic Time Warping (DTW)~\cite{wang2013experimental, jeong2011weighted, berndt1994using}. Others have used distance kernels instead and learned Support Vector Machines (SVMs)~\cite{kampouraki2008heartbeat, bahlmann2002online, shimodaira2001dynamic}, or extracted features and learned linear~\cite{dempster2020rocket} or tree-based classifiers such as eXtreme Gradient Boosting (XGBoost)~\cite{chen2016xgboost}.

Later, deep learning-based approaches are widely adopted because of their ability to learn complex patterns. Convolutional Neural Networks (CNNs) have proven to be successful in learning local patterns in time series data~\cite{wu2022timesnet, franceschi2019unsupervised, zhao2017convolutional}. Similarly, Multilayer Perceptron (MLP) can provide simple but effective time series classifiers~\cite{zhang2022less}. Recurrent Neural Networks (RNNs), such as Long Short-Term Memory (LSTM) effectively handle long sequence modeling~\cite{gu2021efficiently, lai2018modeling, hochreiter1997long}.

More recently, transformer-based models~\cite{vaswani2017attention} have revolutionized the NLP domain, and these models have been adapted to the time series domain~\cite{zhou2022fedformer, wu2021autoformer, zhou2021informer}. The self-attention mechanism in transformers is known for modeling long-range dependencies in sequence data. However, the increasing complexity of these models often comes with larger model sizes and higher computational costs, especially for training.

\section{Datasets}\label{appendix:dataset_details}
We benchmark our model using the following 10 multivariate datasets from the UEA Time Series Classification Archive \citet{bagnall2018uea}. See Table \ref{table:data-characteristics} for their data characteristics.

\begin{table*}[htb]
\centering
\caption{Dataset Characteristics. \textbf{Abbreviations:} EC: EthanolConcentration, FD: FaceDetection, HW: Handwriting, HB: Heartbeat, JV: JapaneseVowels, PEMS: PEMS-SF, SCP1: SelfRegulationSCP1, SCP2: SelfRegulationSCP2, SAD: SpokenArabicDigits, UW: UWaveGestureLibrary}
\label{table:data-characteristics}
\resizebox{\textwidth}{!}{%
\begin{tabular}{@{}lccP{3cm}ccccccc@{}}
\toprule
Characteristic  & EC & FD & HW & HB & JV & PEMS-SF & SCP1 & SCP2 & SAD & UW \\ 
\midrule
Train Size            & 261                  & 5890          & 150         & 204       & 270            & 267     & 268                & 200                & 6599               & 120                 \\
Test Size             & 263                  & 3524          & 850         & 205       &  370            & 173     & 293                & 180                & 2199               & 320                 \\
Number of Dimensions  & 3                    & 144           & 3           & 61        & 12             & 963     & 6                  & 7                  & 13                 & 3                   \\
Series Length         & 1751                 & 62            & 152         & 405       & 29             & 144     & 896                & 1152               & 93                 & 315                 \\
Number of Classes     & 4                    & 2             & 26          & 2         & 9              & 7       & 2                  & 2                  & 10                 & 8                   \\
Type & Spectro & EEG & Motion/Human Activity Recognition & Audio & Audio & Occupancy rate & EEG & EEG & Speech & EEG \\
\bottomrule
\end{tabular}
}
\end{table*}

\subsection{EthanolConcentration}
EthanolConcentration \citep{large2018detecting} comprises raw spectra from water-and-ethanol solutions contained within 44 unique, real whisky bottles, featuring ethanol concentrations of 35\%, 38\%, 40\%, and 45\%. Scotch Whisky regulations require a minimum alcohol content of 40\%, a standard that producers adhere to in order to comply with labeling specifications.  The dataset presents a classification task to identify the ethanol concentration from spectral readings of any given bottle. Each record includes three spectral readings from the same bottle and batch, obtained by positioning the bottle between a light source and a spectroscope. These spectral readings, which cover wavelengths from 226nm to 1101.5nm at a 0.5nm resolution, were recorded over a one-second integration time using a StellarNet BLACKComet-SR spectrometer. The methodology deliberately avoids optimizing for clarity or consistency in the spectral path, aiming to simulate the varied conditions typical of rapid screening tests that may be performed on batches of spirits for quality assurance.

\subsection{FaceDetection}
The FaceDetection dataset originates from a 2014 Kaggle competition \citep{face_detection_dataset}. The challenge involves identifying whether a subject is viewing a picture of a face or a scrambled image using magnetoencephalography (MEG) data, independent of the individual subject. This dataset specifically includes only the training portion from the competition, organized by patient. It comprises data from 10 training subjects (subject01 to subject10) and 6 testing subjects (subject11 to subject16). Each subject has approximately 580 to 590 trials, resulting in a total of 5,890 training trials and 3,524 test trials. Each trial features 1.5 seconds of MEG data, initiated 0.5 seconds before the stimulus is presented, and is associated with a class label—Face (class 1) or Scramble (class 0). The data were down-sampled to 250Hz and subjected to a high-pass filter at 1Hz, producing 62 observations per channel. 

\subsection{Handwriting}
The Handwriting dataset \citep{shokoohi2017generalizing} consists of motion data captured from a smartwatch while subjects wrote the 26 letters of the alphabet. Developed at the University of California, Riverside (UCR), this dataset includes 150 training cases and 850 test cases. It features six dimensions, comprising three accelerometer readings and three gyroscope readings.

\subsection{Heartbeat}
The Heartbeat dataset originates from the PhysioNet/CinC Challenge 2016 \citet{liu2016open} and consists of cardiac sound recordings from a diverse pool of participants, both healthy individuals and patients with cardiac conditions. Recordings were made in various settings, clinical and non-clinical, and captured from multiple body locations including the aortic, pulmonic, tricuspid, and mitral areas, among up to nine potential sites. The dataset categorizes these sounds into two primary classes: normal and abnormal. Normal heart sounds were obtained from healthy subjects, while abnormal sounds were recorded from patients diagnosed with cardiac ailments, predominantly heart valve defects such as mitral valve prolapse, mitral regurgitation, aortic stenosis, and post-valvular surgery conditions, as well as coronary artery disease.

The audio recordings, inclusive of contributions from both children and adults, were uniformly truncated to five seconds. Spectrograms of each truncated audio were generated using a window size of 0.061 seconds with a 70\% overlap. This multivariate dataset is structured with each dimension representing a frequency band derived from the spectrogram. There are 113 instances in the normal class and 296 in the abnormal class.

\subsection{JapaneseVowels} 
The Japanese Vowels dataset \citet{misc_japanese_vowels_128}, sourced from the UCI Machine Learning Repository, comprises recordings from nine male speakers who pronounced the Japanese vowels `a' and `e'. Each utterance was analyzed using a 12-degree linear prediction to extract a 12-dimensional time-series representation, with lengths varying originally from 7 to 29. For consistency, all instances in the dataset have been padded to the maximum length of 29. The objective of the classification task is to identify the speaker; hence each 12-by-29 instance matrix is associated with a single class label, ranging from 1 to 9. This dataset serves as a benchmark for assessing the efficacy of time-series classification models in distinguishing speakers based on LPC cepstrum coefficients obtained from their speech patterns.

The dataset includes a total of 640 time-series instances. A training set consists of 30 utterances per speaker, totaling 270 instances. The test set, however, comprises 370 instances and varies in distribution—ranging from 24 to 88 instances per speaker—owing to external factors such as timing and availability during the experimental setup.

\subsection{PEMS-SF}
The PEMS-SF dataset \citet{cuturi2011fast} contains 15 months of daily data sourced from the California Department of Transportation. This dataset details the occupancy rates, ranging from 0 to 1, across various car lanes on the freeways of the San Francisco Bay Area. The data spans from January 1, 2008, to March 30, 2009, with measurements taken every 10 minutes. Each day is treated as an individual time series with a dimension of 963, corresponding to the number of sensors that consistently functioned throughout the observation period. The length of each time series is 144 data points (6 per hour x 24 hours). The dataset excludes public holidays and two anomalous days (March 8, 2009, and March 9, 2008) when sensors recorded no data between 2:00 and 3:00 AM, resulting in a total of 440 valid time series. The classification task involves identifying the day of the week for each series, labeling them with integers from 1 (Monday) to 7 (Sunday). Each attribute within a record reflects the occupancy rate recorded by a sensor at a specific timestamp throughout the day.

\subsection{SelfRegulationSCP1}
The SelfRegulationSCP1 dataset, sourced from \citet{birbaumer1999spelling}, involves recordings from a healthy subject who was instructed to control a cursor on a screen using cortical potentials. This process was facilitated by tracking the subject's slow cortical potentials (Cz-Mastoids), where cortical positivity resulted in downward cursor movements and cortical negativity caused it to move upward. Each trial, lasting six seconds, was designed to capture these dynamics, with visual feedback provided between the second 2 and 5.5 of the trial. During each trial, a goal was visually indicated at either the top or bottom of the screen starting from 0.5 seconds to the end of the trial, guiding the subject to generate negative or positive potentials correspondingly. The usable data for each trial, however, spans only 3.5 seconds—from the second 2 to 5.5—corresponding to 896 samples per channel given the sampling rate of 256 Hz.

Data capture involved a PsyLab EEG8 amplifier and a PCIM-DAS1602/16 A/D converter, recording over channels positioned according to the 10/20 system. The dataset includes a training set of 268 trials—168 from the first day and 100 from the second, mixed randomly—and 293 test instances, with class labels indicating positivity or negativity.

\subsection{SelfRegulationSCP2}
The SelfRegulationSCP2 dataset \citet{birbaumer1999spelling} includes data from an artificially respirated ALS patient who was tasked with controlling a cursor on a computer screen using cortical potentials. Auditory and visual cues were used to guide the patient, with slow cortical potentials measured at the Cz-Mastoids. A positive potential moved the cursor downward, whereas a negative potential moved it upward. Each trial lasted 8 seconds, with the cursor movement direction (up for negativity, down for positivity) indicated both visually and auditorily from the 0.5 to 7.5 second marks. Auditory instructions were given precisely at the 0.5-second mark, and visual feedback was available from seconds 2 to 6.5. Only the data from this 4.5-second feedback period, translating to 1152 samples per channel at a 256 Hz sampling rate, are used for training and testing.

EEG data were collected from several sites according to the 10/20 system and included channels for detecting vertical eye movements (vEOG). The EEG signals were not corrected for EOG artifacts, providing a raw view of the cortical activity. The dataset comprises 200 trials for training, evenly split between two classes, and an additional 180 trials for testing, recorded on the same day but after the training session data. Each trial spans 7 dimensions and a series length of 1152.

\subsection{Spoken Arabic Digits}
The Spoken Arabic Digits dataset \citet{misc_spoken_arabic_digit_195} consists of 8,800 time series data entries derived from the vocal utterances of 88 native Arabic speakers (44 males and 44 females, aged between 18 and 40). Each dataset entry represents one of ten Arabic digits, spoken ten times by each speaker. The dataset captures 13 Mel Frequency Cepstral Coefficients (MFCCs) for each sound snippet, which are extracted under the following audio processing conditions:
\begin{itemize}
    \item \textbf{Sampling rate:} 11025 Hz
    \item \textbf{Bit depth:} 16 bits
    \item \textbf{Window function:} Hamming
    \item \textbf{Pre-emphasis filter:} $1 - 0.97Z^{-1}$
\end{itemize}

Each line in the database corresponds to one frame of analysis, listing the 13 MFCCs separated by spaces. These coefficients effectively capture the spectral properties essential for recognizing spoken digits. This structured approach facilitates robust time-series analysis for speech recognition tasks involving Arabic numerals.

\subsection{UWaveGestureLibrary}
The UWaveGestureLibrary \citet{liu2009uwave} comprises a set of eight simple gestures, each generated from accelerometer data. The dataset records the X, Y, and Z coordinates corresponding to each gesture's motion. Every time series within this dataset consists of 315 data points. 

\section{Comparison Baselines: Supervised Learning Methods} \label{appendix:supervised_baselines}
To provide a thorough comparison, we evaluate our approach against 21 supervised baseline models for time series classification. These baselines can be categorized into Classical methods, models based on MLPs, RNNs, CNNs, transformers, and LLMs. The details of which are provided below.

\subsection{Classical methods}
\begin{itemize}
\item[1)] \textbf{Dynamic Time Warping (DTW)} \citet{berndt1994using} is a method for measuring similarity between two time series, \(X = (x_1, x_2, \ldots, x_M)\) and \(Y = (y_1, y_2, \ldots, y_N)\), which may differ in length and are sampled at equidistant points in time. DTW identifies the best alignment between these series by minimizing the effects of distortion and shifting in time, allowing for the comparison of similar shapes across different phases.

The core of DTW is the construction of a local cost matrix, \(C \in \mathbb{R}^{M\times N}\), with entries \(c_{i, j} = \|x_i - y_j\|\) for \(i \in [1, M]\) and \(j \in [1, N]\), which represents the pairwise distances between points in the two series. The objective is to find a warping path $p = (p_1, p_2, \ldots, p_L)$ where each \(p_l = (p_i, p_j)\) lies within \([1, M] \times [1, N]\). This path aligns the series by following the route that minimizes cumulative distance, adhering to several constraints: it must start and end at \(p_1 = (1, 1)\) and \(p_L = (M, N)\) (boundary condition), maintain the temporal ordering of points \(m_1 \leq m_2 \leq \ldots \leq m_L\) and \(n_1 \leq n_2 \leq \ldots \leq n_L\) (monotonicity condition), and prevent large temporal jumps (step size condition). 
The optimal warping path is identified through a recursive process aimed at minimizing the total cost associated with \(p\), calculated as \(c_p(X, Y) = \sum_{l=1}^{L} c(x_{m_l}, y_{n_l})\). This path, \(P^*\), where \(c_{P^*} = \min_{p \in P} c_p(X, Y)\), defines the DTW distance, quantifying the similarity between the series. However, the computational cost of this process is \(O(MN)\), where \(M\) and \(N\) are the lengths of the two series, rendering it computationally demanding for large datasets.

For classifying time series, the DTW distance is integrated with the k-nearest neighbors (k-NN) algorithm. This approach computes the DTW distance of a target series to all other series in a training dataset and assigns a classification based on the most common class among the k-nearest neighbors. Thus, DTW effectively accommodates series with time shifts, providing a robust, distance-based method for classifying time series.

\item[2)] \textbf{eXtreme Gradient Boosting (XGBoost)} \citep{chen2016xgboost} is a state-of-the-art machine learning algorithm that primarily uses decision trees as base learners to construct a robust ensemble model. XGBoost sequentially builds a series of weak learners—typically decision trees—and enhances each successive tree by correcting errors made by its predecessors, a technique known as boosting. This process involves an additive model where each new decision tree is improved by leveraging the cumulative knowledge of the trees that came before it, optimizing for maximum information gain at each split using a greedy algorithm. To prevent overfitting, XGBoost incorporates regularization directly into its loss function and employs shrinkage to moderate the learning rate. Additionally, the optimization method, which is gradient-based, minimizes a cost function by iteratively adjusting the model's parameters in response to the gradients of the errors. XGBoost also refines the decision tree construction process by using a Similarity Score and Gain to determine the most effective node splits, further improving the model's accuracy and efficiency.

\item[3)] \textbf{RandOm Convolutional KErnel Transform (ROCKET)} \citep{dempster2020rocket} is a method for time series classification that employs random convolutional kernels to transform series data, which is then used to train a linear classifier. Unlike traditional convolutional neural networks (CNNs) that rely on learned kernels, ROCKET utilizes a broad array of random kernels. Each kernel is uniquely characterized by random properties such as length, weights, biases, dilation, and padding. This configuration forms a single-layer convolutional neural network, where the randomized kernel weights contribute to generating input for a softmax layer, thus optimizing the feature extraction process. ROCKET also efficiently scales for large datasets due to its linear complexity relative to the length of the time series and the number of training samples. The key advantages of ROCKET include:
\begin{itemize}
    \item Number of Kernels: ROCKET utilizes a substantial number of kernels in a single layer. The non-learned nature of these kernels reduces computational costs significantly, enabling the use of numerous kernels without substantial overhead.

    \item Variety of Kernels: Unlike typical CNNs, where kernels might share characteristics, each ROCKET kernel is distinct in its attributes, enhancing the diversity and the ability to detect various patterns.

    \item Kernel Dilation: Dilation in ROCKET is randomly assigned to each kernel, differing from the exponential increase with depth seen in traditional CNNs. This randomness is vital for identifying patterns across different scales and frequencies.

    \item Feature Extraction: Beyond employing global max pooling techniques through the maximum value of feature maps, ROCKET utilizes a novel metric—the proportion of positive values. This metric allows classifiers to assess the prevalence of patterns more accurately within the time series.
\end{itemize}

\end{itemize}

\subsection{MLP-based methods}
\begin{itemize}
\item[4)]  \textbf{LightTS} \citep{zhang2022less} is an MLP-based time series forecasting model that employs simple MLP structures to manage both short-term and long-term temporal dependencies. This model includes two downsampling strategies: interval sampling, which targets long-term dependencies, and continuous sampling, which focuses on short-term local patterns. These strategies are based on the principle that downsampling typically preserves the majority of a time series' crucial information, thus maintaining model efficiency. LightTS utilizes an MLP-based framework on top of these downsampling techniques, enabling effective information exchange among different down-sampled subsequences and time steps. This configuration allows LightTS to adaptively select relevant information for forecasting and to efficiently handle very long input sequences by processing only a fraction of the data after downsampling.

\item[5)] \textbf{DLinear} \citep{zeng2023transformers} is a recent non-transformer model developed in response to the difficulty transformer-based models face in capturing ordering information within time series. DLinear integrates a decomposition scheme similar to those used in Autoformer and FEDformer but relies on linear layers for processing. Initially, it decomposes a raw data input into a trend component using a moving average kernel and a remainder (seasonal) component. Subsequently, two single-layer linear layers are applied independently to each component. The outputs from these layers are then summed to produce the final prediction. This approach allows DLinear to enhance performance over a standard linear model by explicitly handling trends within the data.
\end{itemize}

\subsection{RNN-based models}

\begin{itemize}
\item[6)] \textbf{Long Short-Term Memory (LSTM)} \citet{hochreiter1997long} addresses the vanishing gradient problem that plagues vanilla RNNs in processing longer sequences. This issue arises as the network propagates forward, and the small weight values in the hidden layers are multiplied repeatedly, causing the gradients to diminish rapidly. As a result, the weights in the initial layers become increasingly difficult to train, which impacts the training of subsequent weights, making RNNs challenging to train overall. LSTM mitigates this problem by incorporating a memory cell equipped with various gates that regulate the flow of information into and out of the cell, enabling it to handle long-short term dependencies effectively. An LSTM unit utilizes a cell state and three gates—input, forget, and output—to manage information. Each gate includes a sigmoid layer \(\sigma\) that outputs values between 0 and 1, representing the proportion of information allowed through the gate, and a point-wise multiplication operation. Specifically, the forget gate \(f_t = \sigma(W_f \cdot [h_{t-1}, \, x_t] + b_f)\) determines which information to discard from the previous cell state \(c_{t-1}\) by analyzing the current input \(x_t\) and the previous hidden state \(h_{t-1}\). The input gate \(i_t = \sigma(W_i \cdot [h_{t-1}, \, x_t] + b_i)\) decides which new information to update, and the update to the cell state \(\tilde{c}_t = \tanh(W_c \cdot [h_{t-1}, \, x_t] + b_c)\) is computed. The cell state is then updated to \(c_t = f_t \cdot c_{t-1} + i_t \cdot \tilde{c}_t\). Finally, the output gate \(o_t = \sigma(W_o \cdot [h_{t-1}, \, x_t] + b_o)\) determines what portion of the cell state to output, with the output hidden state given by \(h_t = o_t \cdot \tanh(c_t)\).

\item[7)] \textbf{Long- and Short-term Time-series Network (LSTNet)} \citep{lai2018modeling} incorporates both CNN and RNN components to perform comprehensive time series analysis. The CNN extracts short-term local dependency patterns from multi-dimensional input variables, while the RNN is tasked with capturing complex long-term dependencies in time series trends. To address the issue of scale insensitivity commonly found in neural network models, LSTNet integrates a traditional autoregressive model. Additionally, LSTNet features a Recurrent-skip structure designed to effectively capture very long-term dependence patterns and to facilitate easier optimization, leveraging the periodic properties of the input time series signals. Further enhancing its robustness, LSTNet also employs a traditional autoregressive linear model in parallel with its nonlinear neural network components. This dual approach makes the network particularly adept at managing time series that exhibit significant scale variations.

\item[8)] \textbf{Linear State Space Layer (LSSL)} \citep{gu2021efficiently}, which is part of the structured state space sequence model, introduces a parameterization for state space models (SSM) designed to enhance computational efficiency. LSSL modifies the structured state matrices by decomposing them into a low-rank and a normal term \cite{gu2020hippo, voelker2019legendre}. This modification facilitates the computation of the truncated generating function in frequency space, rather than expanding the standard SSM in coefficient space. Furthermore, LSSL employs the Woodbury identity to adjust the low-rank term, and the normal term is stably diagonalized. This approach simplifies the computations by involving processes associated with a Cauchy kernel \cite{pan2012structured, pan2017fast}, known for its stability in theoretical contexts. These modifications allow LSSL to efficiently manage both computational and memory resources.
\end{itemize}

\subsection{CNN-based models}
\begin{itemize}
\item[9)] \textbf{Temporal Convolutional Network (TCN)} \citep{bai2018empirical, franceschi2019unsupervised} is a CNN-based architecture designed to capture extended historical data with long memory capabilities and requires minimal tuning in practice. TCNs utilize dilated causal convolutions to ensure that predictions do not prematurely incorporate future data. The dilations significantly expand the network's receptive field, enabling it to cover a broader range of historical context. Furthermore, TCNs integrate residual connections to facilitate the effective training of deeper models.

A TCN model comprises a series of \(n\) TCN residual blocks, where \(n\) is a hyperparameter. Each block contains two dilated causal convolutional layers, which are fully convolutional to ensure that the output size is consistent with the input size. These causal convolutions guarantee that the output at any given time \(t\) depends only on inputs from time \(t\) and earlier. The convolutional layers are applied with a stride of 1, and padding adjustments maintain the convolutional nature of the network. Each convolutional layer applies a dilation factor \(d\), typically set as \(d = 2^i\) for the \(i\)-th block, to exponentially increase the receptive field as the network deepens. Mathematically, a convolution with dilation factor \(d\) on an element \(x\) of a 1D input \(g\) with a filter \(f\) of length \(k\) is computed as \((g \, *_d \, f) (x) = \sum_{j=0}^{k-1}f(j)\cdot g(x-d\cdot j)\).

Following the convolutional layers, the sequence of operations includes weight normalization \cite{salimans2016weight}, ReLU activation, and a dropout layer. Weight normalization improves gradient conditioning and accelerates convergence by reparameterizing each weight vector \(\textbf{w}\) as \(\textbf{w} = \frac{g}{||\textbf{v}||} \textbf{v}\), where \(\textbf{v}\) has a fixed norm and \(g\) is a scalar. This process decouples the magnitude of the weight vector from its direction, enhancing network optimizability. Each block concludes with an element-wise addition of the block’s input and the estimated residual mapping, followed by a ReLU activation. Dimension alignment for this addition is achieved with a \(1 \times 1\) convolution. 

\item[10)] \textbf{TimesNet} \citep{wu2022timesnet} is a method that overcomes the limitations of traditional 1D time series representation by transforming these series into 2D tensors organized across multiple periods. It captures short-term intraperiod variations and long-term interperiod trends by embedding them into the columns and rows of the 2D tensors, respectively. This transformation allows for more efficient modeling of temporal variations using 2D convolutional kernels, thereby extending the analysis into a more comprehensive 2D space. TimesNet ensures simultaneous representation of both intraperiod and interperiod variations, with modules specifically tailored to emphasize the unique temporal patterns of each period.

The central component of TimesNet, the TimesBlock, is a versatile and adaptive structure designed to detect multiperiodicity and extract complex temporal patterns from these 2D tensors. Its parameter-efficient Inception-block \citep{szegedy2015going} based architecture boosts the model's analytical capabilities, facilitating a detailed examination of distinct temporal variations associated with different periods. This approach allows TimesNet to transcend the constraints of 1D representations, enabling a unified and thorough analysis of temporal variations.
\end{itemize}

\subsection{Transformer-based models}
\begin{itemize}

\item[11)] \textbf{Transformer} \citep{vaswani2017attention} comprises a dual-component architecture with both an encoder and a decoder, each containing stacked self-attention and point-wise, fully connected layers. Each component consists of six identical layers. In the encoder, each layer includes a multi-head self-attention mechanism and a position-wise fully connected feed-forward network, complemented by a residual connection and layer normalization. The decoder replicates this configuration but adds a third sub-layer for multi-head attention on the encoder's output. It also modifies its self-attention mechanism to prevent forward-looking attention, thus preserving the model's auto-regressive properties for sequential generation.

The Transformer utilizes multi-head attention to enable nuanced interactions between its encoder and decoder and to facilitate detailed processing across different positions in the sequence. This attention mechanism projects queries, keys, and values through multiple linear transformations, enabling diverse representation and integration of information across subspaces. It is applied in three distinct forms: encoder-decoder attention allows decoder queries to attend to all positions in the encoder output; self-attention within the encoder lets each position process information from all preceding positions; and self-attention within the decoder restricts attention to prevent future positions from influencing the sequence, maintaining sequential integrity. This structured approach helps the Transformer effectively manage and process long sequence dependencies, making it adaptable for a broad range of sequence-based applications.

\item[12)] \textbf{Reformer} \citep{kitaev2020reformer} enhances the efficiency of the transformer model with two significant modifications, making it more suitable for processing long sequences. Firstly, it replaces the traditional dot-product attention with a locality-sensitive hashing mechanism. This change reduces the computational complexity from \(O(L^2)\) to \(O(L \log L)\), where \(L\) is the sequence length. Secondly, Reformer employs reversible residual layers, which regenerate the activations of any layer from the subsequent layer's activations using only the model parameters. This approach eliminates the need for storing multiple copies of activations for each layer, substantially reducing memory usage. The reversible layers, introduced in \cite{gomez2017reversible}, require only a single set of activations to be stored for the entire model, significantly reducing the memory cost usually multiplied by the number of layers (\(N\)). Moreover, the Reformer processes activations in chunks within feed-forward layers, further decreasing memory demands. These adjustments, along with the use of locality-sensitive hashing for attention computation, not only minimize memory and computational overhead but also maintain the model's performance on par with the traditional transformer model for long sequences.

\item[13)] \textbf{Informer} \citep{zhou2021informer} is a transformer-based model designed specifically to tackle challenges in long sequence time-series forecasting, such as quadratic time complexity, high memory usage, and constraints of the traditional encoder-decoder architecture. The Informer introduces several modifications to enhance efficiency and effectiveness in processing long sequences:
\begin{itemize}
    \item ProbSparse Self-Attention Mechanism: This mechanism replaces the conventional self-attention in Transformers. It is engineered to achieve \(O(L \log L)\) in both time complexity and memory usage, efficiently managing dependency alignments without compromising performance.
    
    \item Self-Attention Distilling: This process improves attention management by concentrating on dominant attention scores and halving the input size for each cascading layer. This method effectively manages extremely long input sequences and significantly lowers the space complexity to \(O((2 - \epsilon) L \log L)\).
    
    \item Generative Style Decoder: Diverging from the typical step-by-step decoding process, the generative style decoder in Informer predicts long time-series sequences in a single forward operation. This approach accelerates inference for long-sequence predictions and reduces the propagation of cumulative errors during the inference phase.
\end{itemize}
The Informer leverages these enhancements—ProbSparse self-attention for efficient processing, self-attention distilling to focus on important attention scores, and a generative style decoder for rapid sequence generation—to improve its performance in forecasting long sequences and capturing long-range dependencies between extensive time-series inputs and outputs.

\item[14)]  \textbf{Pyraformer} \citep{liu2021pyraformer} is a transformer-based model that employs a pyramidal attention module (PAM) for efficient management of time series data. This module uses inter-scale and intra-scale connections to summarize features at different resolutions and capture temporal dependencies across various ranges. The design integrates a tree structure for inter-scale connections and neighboring intra-scale connections to achieve a multi-resolution representation of time series. This architecture ensures that Pyraformer scales linearly with the input series length, optimizing computational efficiency while maintaining a constant signal path length relative to the sequence length \(L\), and keeping both time and space complexity linear with \(L\).

Pyraformer operates by first embedding observed data, covariates, and positions in a manner similar to the Informer \citep{zhou2021informer}. It then constructs a multi-resolution \(C\)-ary tree through a coarser-scale construction module (CSCM), where each coarser scale node aggregates information from \(C\) finer scale nodes. This structure allows Pyraformer to model temporal dependencies efficiently across different scales through sparse intra-scale connections, thus reducing computational overhead. Depending on the specific needs of different downstream tasks, Pyraformer adapts its output structure to effectively meet the requirements of diverse time series analyses.

\item[15)]  \textbf{Autoformer} \citep{wu2021autoformer} is a transformer-based model that incorporates time series decomposition, drawing inspiration from classical time series analysis methods. Unlike traditional transformers that rely on self-attention mechanisms to capture long-range dependencies, Autoformer introduces an auto-correlation mechanism as an alternative. This change addresses the issues that traditional models face with intricate temporal patterns and long-term forecasting, where identifying reliable dependencies can be challenging. To enhance efficiency in handling long series, traditional transformers sometimes use sparse versions of self-attention, which can limit the effective use of information.

Autoformer integrates decomposition blocks directly into its structure, moving away from the typical preprocessing approach of series decomposition. These blocks are specifically designed to progressively isolate long-term trends from the data during the forecasting process, allowing the model to refine and decompose the data iteratively. This setup improves the handling of complex time series. The auto-correlation mechanism, inspired by stochastic process theory and based on the periodicity of the series, focuses on identifying dependencies and aggregating representation at the sub-series level, effectively capturing and utilizing similarities derived from underlying periodic patterns.

The architecture of Autoformer adheres to a residual and encoder-decoder framework. The encoder eliminates long-term trend-cyclical components through the series decomposition blocks and focuses on modeling seasonal patterns. In contrast, the decoder accumulates the trend component extracted from the hidden variables, using past seasonal information to enhance forecasting accuracy. Additionally, Autoformer incorporates a moving average within its decomposition blocks to smooth out periodic fluctuations and highlight long-term trends, thereby facilitating a more targeted analysis of stable trend components within the time series.

\item[16)] \textbf{Non-stationary Transformer} \citep{liu2022non} addresses the challenges posed by non-stationary real-world data, where the joint distribution changes over time, often leading to the degradation of transformer performance. Non-stationary Transformer consists of two interdependent modules: series stationarization and de-stationary attention. Series stationarization normalizes the input data to unify its statistical properties, enhancing predictability, and adjusts the output to restore the original statistics. This module utilizes a straightforward normalization approach without additional parameters. De-stationary attention, on the other hand, aims to counteract the potential over-normalization by reintroducing the intrinsic non-stationary characteristics of the data into the model's temporal dependencies. The de-stationary attention module approximates how attention mechanisms would function on unnormalized data and integrates these insights back into the model to maintain crucial temporal dynamics. This setup allows stationformer to balance the predictability benefits gained from normalized data with the rich, detailed patterns inherent in the raw non-stationary series.

Structurally, Non-stationary Transformer adapts the traditional encoder-decoder setup. The encoder extracts information from past observations, while the decoder aggregates this information to refine predictions. The framework modifies the standard transformer by applying series stationarization to both the input and output of the model, and replaces the conventional self-attention mechanism with de-stationary attention. This adaptation aims to enhance the model’s ability to predict non-stationary series by effectively managing the challenges associated with data variability over time.

\item[17)]  \textbf{Frequency Enhanced Decomposed Transformer (FEDformer)} \citep{zhou2022fedformer} is a transformer-based method that incorporates a time series decomposition scheme to address the limitations of traditional transformers, particularly their high computational demands and challenges in capturing global time series trends. By combining transformers with the seasonal-trend decomposition method, FEDformer aims to separate the broad trends from more detailed fluctuations in time series data. This allows the transformer component to focus on more granular details while the decomposition handles the overall profile of the series.

The architecture of FEDformer includes specialized blocks such as Fourier-enhanced and Wavelet-enhanced blocks within the transformer framework. These blocks serve as substitutes for the conventional self-attention and cross-attention mechanisms, and they enable the model to analyze important structures through frequency domain mapping. FEDformer employs a selective approach to incorporating Fourier components, which helps keep the computational complexity and memory usage linear in relation to the length of the time series. Specifically, FEDformer is structured as a deep decomposition architecture that integrates Frequency Enhanced Block (FEB), Frequency Enhanced Attention (FEA) connecting the encoder and decoder, and the Mixture Of Experts Decomposition block (MOEDecomp). This setup leverages both seasonal-trend decomposition and distribution analysis to facilitate the processing of time series data.

\item[18)] \textbf{ETSformer} \citep{woo2022etsformer} is a transformer-based architecture tailored for time series forecasting, integrating exponential smoothing techniques. ETSformer uses two novel mechanisms, Exponential Smoothing Attention (ESA) and Frequency Attention (FA), which are designed to replace the traditional self-attention mechanism in standard transformers. ESA utilizes attention scores based on relative time lags, enabling efficient handling of growth components with a computational complexity of \(O(L\log L)\) for a length-\(L\) lookback window. Similarly, FA employs Fourier transformations to identify dominant seasonal patterns, selecting bases with the highest amplitudes in the frequency domain to achieve the same level of complexity.

ETSformer is structured with modular decomposition blocks that allow it to dissect time-series data into distinct, interpretable components such as level, growth, and seasonality. This design facilitates layer-wise decomposition of the time series into these components, enhancing the model's ability to capture and represent complex temporal dynamics. The architecture systematically extracts latent growth and seasonal patterns through a deep, multi-layered approach, where each layer progressively refines the extraction of these temporal features. The final forecast generated by ETSformer integrates these decomposed elements—level, trend, and seasonality—into a cohesive output that is both practical and interpretable for human analysts. By emphasizing recent observations, the model aligns with the principles of exponential smoothing, ensuring that more recent trends carry greater weight in the forecast. 

\item[19)] \textbf{Flowformer} \citep{wu2022flowformer} modifies the traditional transformer architecture by integrating flow network theory to tackle the scalability issues typical of standard transformers. The conventional attention mechanism in transformers is known for its quadratic complexity, which limits their ability to process a large number of tokens and scale to larger models effectively. To address this, Flowformer introduces the Flow-Attention mechanism, grounded in the principles of flow conservation. This mechanism reimagines attention as information flowing from sources (values) to sinks (results) via learned flow capacities (attentions), aiming to achieve linear complexity.

The Flow-Attention mechanism manages the flow of information by regulating incoming flow at sinks to initiate source competition and outgoing flow at sources for sink allocation. This management of flow helps in aggregating relevant information without relying on specific inductive biases and aims to prevent the common issue of degenerated attentions found in typical attention mechanisms. Flowformer embeds flow conservation within its attention mechanism to streamline the process of information aggregation and refinement. By integrating these elements, Flowformer seeks to provide an alternative approach that could potentially handle large datasets and complex time series data more efficiently than traditional transformers, without the computational complexity typically associated with these models.

\item[20)] \textbf{Patch Time Series Transformer (PatchTST)} \cite{nietime} is a transformer-based model for multivariate time series forecasting. It is built on two key principles: (i) segmentation of time series into subseries-level patches, which serve as input tokens to the transformer, and (ii) channel-independence, where each channel contains a single univariate time series that shares the same embedding and transformer weights across all series. The patching mechanism provides three main advantages: it retains local semantic information in embeddings, significantly reduces the computational and memory complexity of attention maps for a given look-back window, and enables the model to attend to longer historical dependencies. The model operates as follows: multivariate time series data is divided into independent channels, each processed separately while sharing the same transformer backbone. Each univariate time series undergoes instance normalization before being segmented into patches, which then serve as input tokens for the transformer. PatchTST employs masked self-supervised learning, where patches are randomly selected and set to zero, requiring the model to reconstruct the missing values. The training objective minimizes Mean Squared Error (MSE) loss between the predicted and ground truth values, demonstrating the effectiveness of transformers when time series data is segmented into patches at the subseries level. For downstream time series tasks, the same transformer encoder is extended and trained with a linear layer for task-specific predictions.
\end{itemize}

\subsection{LLM-based models}
\begin{itemize}
    \item[21)] \textbf{OneFitsAll} \citep{zhou2024one} employs pre-trained language and computer vision models, developed from billions of tokens, for time series analysis. This approach, termed the Frozen Pretrained Transformer (FPT), retains the original self-attention and feedforward (FFN) layers of the pre-trained models, which hold the majority of the learned knowledge. The model is fine-tuned for various time series classification tasks, with adjustments made only to the positional embeddings and layer normalization layers to adapt to specific downstream tasks. Additionally, to more effectively manage local semantic information, OneFitsAll incorporates a patching technique as described by \cite{nietime}. This method aggregates adjacent time steps into a single patch-based token, thereby increasing the historical time horizon that can be processed by the model without increasing the token length, reducing information redundancy within transformer models.
\end{itemize}

\section{Adapting Supervised Forecasting Baseline Models for Classification} \label{adapting_forecasting_to_classification}
It is important to note that some of our supervised learning-based baselines were adapted from forecasting to classification tasks without altering the core design of the models. We employ a lightweight multi-layer perceptron (MLP)-based projection layer on top of the model's encoded features (originally designed for forecasting) to map them to classification labels. The parameters of this layer are learned during training, enabling the model to adapt effectively for classification tasks. This approach aligns with methods used in prior works such as OneFitsAll \cite{zhou2024one} and TimesNet \cite{wu2022timesnet} when adapting forecasting models for multivariate classification.


\section{Comparison Baselines: Unsupervised Representation Learning Methods} 
\label{appendix:unsupervised_baselines}
\subsection{CNN-based models}
\begin{itemize}
\item[1)] \textbf{T-Loss} \cite{franceschi2019unsupervised} is an unsupervised method for learning general-purpose representations of multivariate time series while handling variations in length. It trains a scalable encoder, structured as a deep convolutional neural network with dilated convolutions, to produce fixed-length vector representations regardless of input length. The loss function is designed as a triplet loss with time-based negative sampling, leveraging the encoder's ability to process time series of unequal lengths. The objective is to ensure that similar time series obtain similar representations without requiring supervision to learn such similarity.

\item[2)]
\textbf{Temporal Neighborhood Coding (TNC)} \cite{tonekaboniunsupervised} is a self-supervised approach that leverages the local smoothness of signals to define temporal neighborhoods and learn generalizable representations for time series. It builds on the smoothness of the generative process of signals to capture meaningful representations for time series windows by ensuring that, in the representation space, signals close in time are distinguishable from those farther apart. In other words, temporal proximity remains identifiable in the encoding space. For each sample window, a neighborhood distribution is first defined. The encoder learns the distribution of windows in the representation space, and samples from these distributions are fed into a discriminator, which predicts the probability of the windows belonging to the same neighborhood. TNC is a general framework, making it agnostic to both the nature of the time series and the architecture of the encoder. The encoder can be any parametric model suited to the signal properties. For the discriminator, a simple multi-headed binary classifier is used.

\item[3)] \textbf{TS2Vec} \cite{yue2022ts2vec} is a universal contrastive learning framework for time series representation learning across all semantic levels. It performs hierarchical contrastive learning over augmented context views, enabling robust contextual representations for each timestamp while capturing multi-resolution temporal dependencies.
To learn representations at different granularity, TS2Vec applies a simple aggregation mechanism. The representation of an arbitrary sub-sequence is obtained via max pooling over the corresponding timestamps, allowing the model to generate fine-grained representations at any resolution. The framework hierarchically discriminates positive and negative samples across both instance-wise and temporal dimensions, reinforcing the ability to capture contextual relationships in time series data.

The contrastive objective is based on augmented context views, ensuring that representations of the same sub-series in different augmented contexts remain consistent. To achieve this, two overlapping subseries are randomly sampled from an input time series, and consistency of contextual representations is encouraged on the common segment. The encoder is optimized jointly with temporal and instance-wise contrastive loss, with the total loss summed across multiple scales in a hierarchical manner.
The encoder consists of three key components: an input projection layer, a timestamp masking module, and a dilated CNN module. The input projection layer maps observations at each timestamp into a high-dimensional latent space via a fully connected layer. The timestamp masking module then masks latent vectors at randomly selected timestamps to generate an augmented context view. Importantly, masking is applied at the latent vector level rather than raw input values, as time series values may be unbounded, making it impractical to use a special token for raw data.
\end{itemize}

\subsection{Transformer-based models}
\begin{itemize}
\item[4)] \textbf{Time-Series Representation Learning via Temporal and Contextual Contrasting (TS-TCC)} \cite{eldele2021time} is an unsupervised framework for learning time-series representations from unlabeled data using contrastive learning. It transforms raw time-series data into two different but correlated views through weak and strong augmentations. TS-TCC consists of two key modules: a temporal contrasting module and a contextual contrasting module. The temporal contrasting module enhances robustness by enforcing a cross-view prediction task, where past latent features from one augmentation predict the future of another. This design encourages the model to learn transformation-invariant representations while handling perturbations from different timesteps and augmentations. The contextual contrasting module further refines representation learning by maximizing similarity among different contexts of the same sample while minimizing similarity across samples. By building on the representations learned from temporal contrasting, this module enhances the model’s ability to capture discriminative features. TS-TCC employs simple yet effective augmentations applicable to any time-series data, ensuring adaptability across diverse datasets.

\item[5)] \textbf{Time Series Transformer (TST)} \cite{zerveas2021transformer} is a transformer-based framework for unsupervised representation learning of multivariate time series. It employs a masked Mean Squared Error (MSE) loss to train a transformer model, extracting dense vector representations through an autoregressive input denoising objective. Specifically, parts of the input are masked (set to zero), and the model is trained to predict the missing values. Since transformers are inherently insensitive to input order, TST incorporates positional encoding to capture the sequential nature of time series data. Only the predictions on the masked values contribute to the MSE loss. The pre-trained model can be applied to various downstream tasks, including classification. For classification, the final representation vectors from all time steps are concatenated into a single vector, which serves as input to a linear output layer with parameters corresponding to the number of classes.

\item[6)] \textbf{MOMENT} \cite{goswamimoment} is an open-source foundation model for general-purpose time series analysis. It follows a masked time series modeling approach, where the goal is to reconstruct masked input time series using a lightweight reconstruction head. First, a univariate time series is segmented into disjoint fixed-length sub-sequences (patches). Before patching, reversible instance normalization \cite{kim2021reversible} is applied to the time series. Each patch is then mapped to a embedding, using a trainable linear projection if all time steps are observed, and a designated learnable mask embedding if some patches are masked. During pretraining, patches are randomly masked by replacing their patch embeddings with a special learnable mask embedding. These patch embeddings are then fed into a transformer encoder to learn patch representations. The transformed patch embeddings are used to reconstruct both masked and unmasked time series patches through a lightweight reconstruction head. Pretraining ensures that embeddings and high-level representations of the time series are effectively learned, with the objective of minimizing the masked reconstruction error (Mean Squared Error) between the ground truth and predicted patches. 

To process multivariate time series, MOMENT operates independently on each channel, aligning with recent studies \cite{nietime, zhou2024one} that support channel-wise modeling as an effective strategy. Unlike traditional architectures that use a decoder of similar size to the encoder, MOMENT employs a lightweight reconstruction head, enabling task-specific fine-tuning with minimal trainable parameters while preserving the encoder’s high-level learned features. For classification, MOMENT operates in a zero-shot setting by replacing its reconstruction head with a linear classification head, mapping patch representations to logits corresponding to class labels.
\end{itemize}

\section{Adapting Unsupervised Representation Learning Models for Classification}\label{appendix_adapting_unsupervised_to_classification}

There are two approaches to adapting unsupervised representation learning baseline models for classification:

\begin{itemize}
    \item \textbf{Two-stage approach:} First, sequence-level representations for each time series are obtained without access to labels, using the unsupervised representation learning model. In the second stage, a machine learning classifier (e.g., a Support Vector Machine with an RBF kernel) is trained on these representations with labeled data. The classifier applies a softmax function to generate a probability distribution over classes, with cross-entropy loss computed against the categorical ground truth labels. This approach is used for all unsupervised baselines except MOMENT, specifically for T-Loss, TNC, TS2Vec, TS-TCC, and TST.
    
    \item \textbf{Direct adaptation approach:} Instead of a two-stage process, the reconstruction head in the unsupervised representation learning model is replaced with a linear classification head that maps patch representations to logits corresponding to class labels. MOMENT adopts this approach by replacing its reconstruction head with a linear head that outputs logits equal to the number of classes.
\end{itemize}


\section{Reproduction Details for Baselines} \label{appendix_reproduce_baselines}
Table \ref{appendix_table:baseline_code_reference} lists the original code sources for each baseline model. All models, except MOMENT, were trained on each dataset individually—either with labels for supervised learning methods or without labels for unsupervised representation learning methods. We collect supervised baseline model results from \cite{zhou2024one} and \cite{wu2022timesnet}, except for PatchTST, which is based on our own reproduction. Additionally, we gather unsupervised representation learning baseline results from \cite{goswamimoment} and \cite{yue2022ts2vec}.
\begin{table*}[htb]
\caption{Code source for each baseline.}
\label{appendix_table:baseline_code_reference}
\centering
\resizebox{\linewidth}{!}{%
\begin{tabular}{@{}l|l@{}}
\toprule
\textbf{Model} & \textbf{Source} \\ 
\midrule
\multicolumn{2}{l}{\cellcolor{pale4}\textit{Supervised Learning Methods}}\\
\midrule
DTW & \url{https://github.com/markdregan/K-Nearest-Neighbors-with-Dynamic-Time-Warping/tree/master}\\
XGBoost & \url{https://github.com/dmlc/xgboost} \\
ROCKET & \url{https://github.com/angus924/rocket/tree/master} \\
LightTS & \url{https://github.com/thuml/Time-Series-Library/blob/main/models/LightTS.py} \\
DLinear & \url{https://github.com/thuml/Time-Series-Library/blob/main/models/DLinear.py} \\
LSTNet & \url{https://github.com/laiguokun/LSTNet} \\
LSSL & \url{https://github.com/state-spaces/s4} \\
TCN & \url{https://github.com/White-Link/UnsupervisedScalableRepresentationLearningTimeSeries} \\
TimesNet & \url{https://github.com/thuml/Time-Series-Library/blob/main/models/TimesNet.py} \\
Transformer & \url{https://github.com/thuml/Time-Series-Library/blob/main/models/Transformer.py} \\
Reformer & \url{https://github.com/thuml/Time-Series-Library/blob/main/models/Reformer.py} \\
Informer & \url{https://github.com/thuml/Time-Series-Library/blob/main/models/Informer.py} \\
Pyraformer & \url{https://github.com/thuml/Time-Series-Library/blob/main/models/Pyraformer.py} \\
Autoformer & \url{https://github.com/thuml/Time-Series-Library/blob/main/models/Autoformer.py} \\
Non-stationary Transformer & \url{https://github.com/thuml/Time-Series-Library/blob/main/models/Nonstationary_Transformer.py} \\
FEDformer & \url{https://github.com/thuml/Time-Series-Library/blob/main/models/FEDformer.py} \\
ETSformer & \url{https://github.com/thuml/Time-Series-Library/blob/main/models/ETSformer.py} \\
Flowformer & \url{https://github.com/thuml/Flowformer/tree/main/Flowformer_TimeSeries} \\
PatchTST & \url{https://github.com/yuqinie98/PatchTST} \\
One Fits All & \url{https://github.com/DAMO-DI-ML/NeurIPS2023-One-Fits-All} \\
\midrule
\multicolumn{2}{l}{\cellcolor{pale4}\textit{Unsupervised Representation Learning Methods}}\\
\midrule
T-Loss & \url{https://github.com/White-Link/UnsupervisedScalableRepresentationLearningTimeSeries} \\
TNC & \url{https://github.com/sanatonek/TNC_representation_learning} \\
TS2Vec & \url{https://github.com/zhihanyue/ts2vec} \\
TS-TCC & \url{https://github.com/emadeldeen24/TS-TCC} \\
TST & \url{https://github.com/gzerveas/mvts_transformer} \\
MOMENT & \url{https://github.com/moment-timeseries-foundation-model/moment} 
\\
\bottomrule
\end{tabular}
}
\end{table*}

\section{Implementation Details for LETS-C} \label{appendix:implementation_details}
The experiments were conducted on a Linux machine equipped with an NVIDIA T4 Tensor Core GPU with 16GB of memory. We utilized the PyTorch v2.4.0 deep learning platform running on Python 3.11 for all models. 
We set a fixed random seed for all experiments to ensure reproducibility.
All configurations employed the RAdam optimizer \citet{liu2019variance}, with its default hyperparameters settings \((\beta_1, \beta_2) = (0.9, 0.999)\).
Further, we explored \letsc{}'s performance across three different text embedding models: \texttt{e5-mistral-7b-instruct} \cite{wang-etal-2024-improving-text}, \texttt{gte-large-en-v1.5} \cite{li2023towards}, and \texttt{nomic-embed-text-v1} \cite{nussbaum2024nomic}, in addition to \texttt{text-embedding-3-large} in Section \ref{sec:additional_results}. 
Exploratory hyperparameter optimization was conducted, revealing that 1-3 1D convolutional layers and 1-2 linear layers are optimal for the performance of \letsc{} across all datasets. 
For a detailed description of the hyperparameters, see Appendix Section \ref{appendix:hyperparameter_details}. For tokenization, we found that maintaining a precision of one decimal place optimizes performance (see Appendix Section \ref{appendix:prec_analysis} for details), thus we adopted that precision throughout the experiments.


\section{Hyperparameter Settings for LETS-C} \label{appendix:hyperparameter_details}
\subsection{Hyperparameter Search Space}
Our hyperparameter search for the lightweight classification head included determining the optimal number of convolutional layers, whether to use dropout or pooling within the convolutional blocks, the number of dense layers, the type of activation function for each layer, and whether to employ batch normalization. We explored 1 to 5 convolutional layers and 1 to 4 linear layers. We also tested various learning rates, ranging from 0.001 to 0.05, different activation functions such as ReLU, GELU, and tanh, and truncation lengths for the text-embedding-large from 64 to 1024.

\subsection{Optimal Hyperparameters}
Table \ref{table:hyperparameter} presents the optimal configurations for our experiments across various datasets. All configurations employed tokenization with a precision of one decimal place and utilized the \texttt{text-embedding-3-large} embeddings. It is crucial to note that the count of linear layers includes the output layer; therefore, a configuration with one linear layer means that this single layer functions as the output layer within the model architecture. Furthermore, all convolutional layers utilized a kernel size of 3.
Additionally, each configuration used the RAdam optimizer \citet{liu2019variance} with its default hyperparameter settings \((\beta_1, \beta_2) = (0.9, 0.999)\).

\begin{table*}[htb]
\caption{Optimal hyperparameter configuration for LETS-C.}
\label{table:hyperparameter}
\centering
\resizebox{0.85\linewidth}{!}{%
\begin{tabular}{@{}l|P{2cm}P{2cm}P{1cm}P{1.5cm}|P{1.5cm}P{1cm}@{}}
\toprule
\multicolumn{7}{l}{\bf LETS-C}\\
\midrule
\multirow{3}{2cm}{Dataset/Configuration} & \multicolumn{4}{c|}{Model Hyperparameter} & \multicolumn{2}{c}{Training Process} \\
\cmidrule{2-7}
& Embedding dimension & Conv layers & Linear layers & Activation & Learning rate & Batch size\\ \midrule
EthanolConcentration & 64 & 2 & 1 & tanh & 0.007 & 64 \\
FaceDetection & 64 & 1 & 1 & tanh & 0.007 & 128 \\
Handwriting & 16 & 1 & 2 & tanh & 0.007 & 64 \\
Heartbeat & 512 & 1 & 1 & tanh & 0.007 & 64 \\
Japanese Vowels & 512 & 1 & 2 & GELU & 0.007 &  64 \\
PEMS-SF & 1024  & 3 & 1 & tanh & 0.007 & 64 \\
Self-Regulation SCP1 & 64 & 1 & 1 & tanh & 0.007 & 64 \\
Self-Regulation SCP2 & 1024  & 2 & 1 & GELU & 0.007 & 64 \\
Spoken Arabic Digits & 512 & 1 & 1 & tanh & 0.001 & 64 \\
UWave Gesture Library & 1024 & 1 & 1 & tanh & 0.001 & 64  \\
\bottomrule
\end{tabular}
}
\end{table*}

\section{Model Performance}
\begin{figure*}[htb] 
    \centering
    \includegraphics[scale = 0.7]{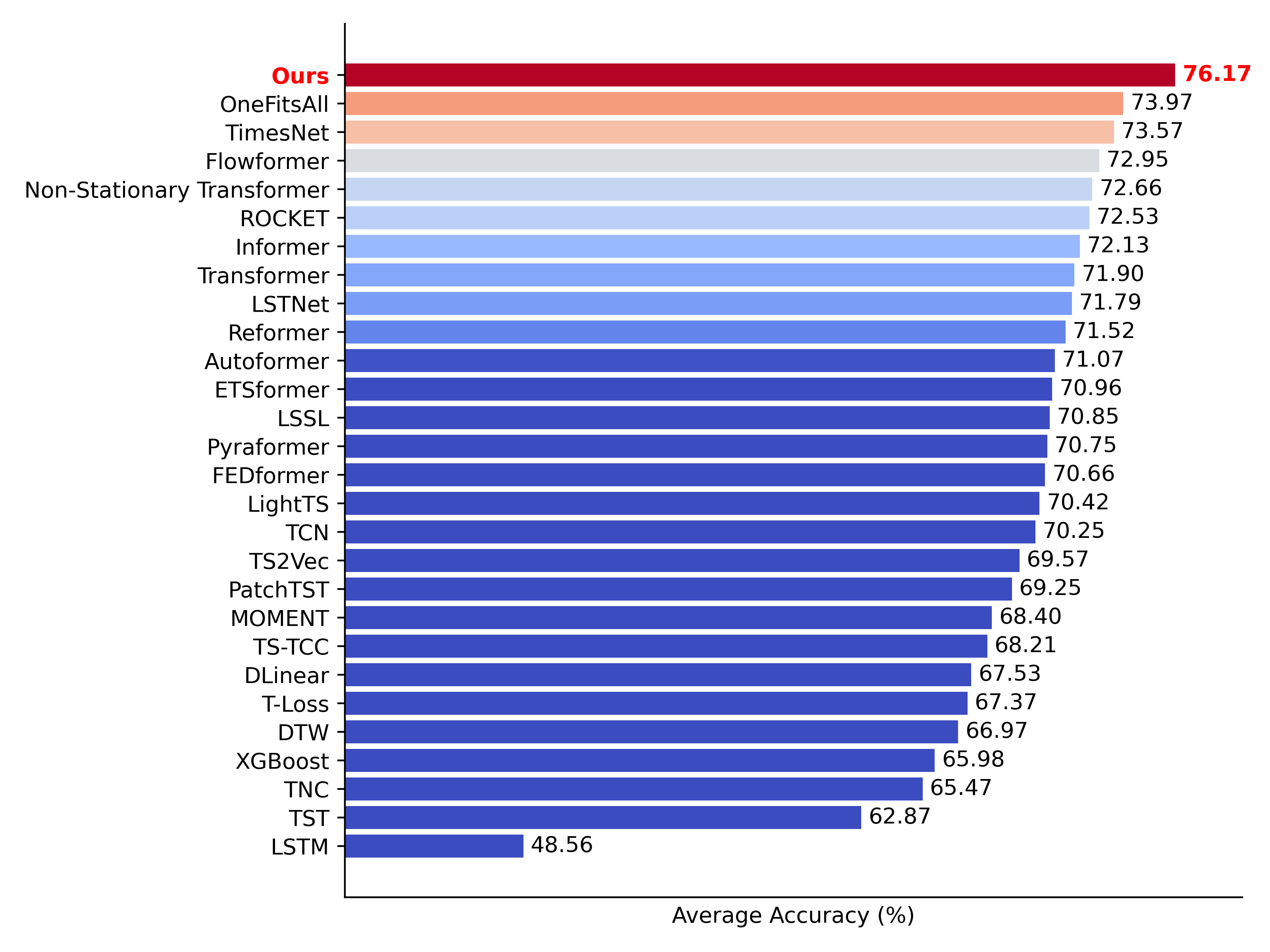}
    \caption{Model comparison based on classification accuracy, averaged across all datasets listed in Table \ref{table:data-characteristics}. For detailed results, refer to Table \ref{table:accuracy} in the main paper.}
    \label{fig:barplot_accuracy}
\end{figure*}

Figure \ref{fig:barplot_accuracy} displays a comparison of models based on the average classification accuracy across all datasets, as summarized in Table \ref{table:data-characteristics}.

\section{Monetary Costs Analysis} \label{sec:monetary_costs}
Our approach incurs monetary costs when utilizing OpenAI's \texttt{text-embedding-3-large} model \cite{openai_large_embeddings}. 
To compute the cost of generating text embeddings for our datasets, we followed the pricing structure of \$0.130 per million tokens. Each time series sample, including all time steps, was embedded separately for each dimension. To estimate the number of tokens per sample, we considered the numerical format of the time series values, where each time step consists of a three-digit number. These digits are separated by spaces, and consecutive time steps are delimited by commas, resulting in an estimated six tokens per time step. The total token count for each dataset was then computed as:
\begin{align*}
\text{Total Tokens} &= 
\text{Total Samples} \times \text{Num. Dimensions} \\
&\quad \times \text{Tokens per Series},
\end{align*}
where the estimated tokens per series were derived from the series length multiplied by six. The dataset characteristics, including train and test sizes, number of dimensions, and series lengths, are detailed in Table~\ref{table:data-characteristics}. Using this approach, the total number of tokens across all datasets was approximately 1.05 billion, leading to an estimated embedding cost of \$137.07. It is important to note that API costs may fluctuate over time, leading to potentially different or outdated values. Furthermore, cost remains a common limitation for all LLM-based approaches that rely on closed-source paid models.


\section{Assessing Text Embeddings with Cosine Similarity}\label{appendix:cosine_similarity}
\begin{figure*}[htb]
    \centering
    \includegraphics[scale = 0.27]{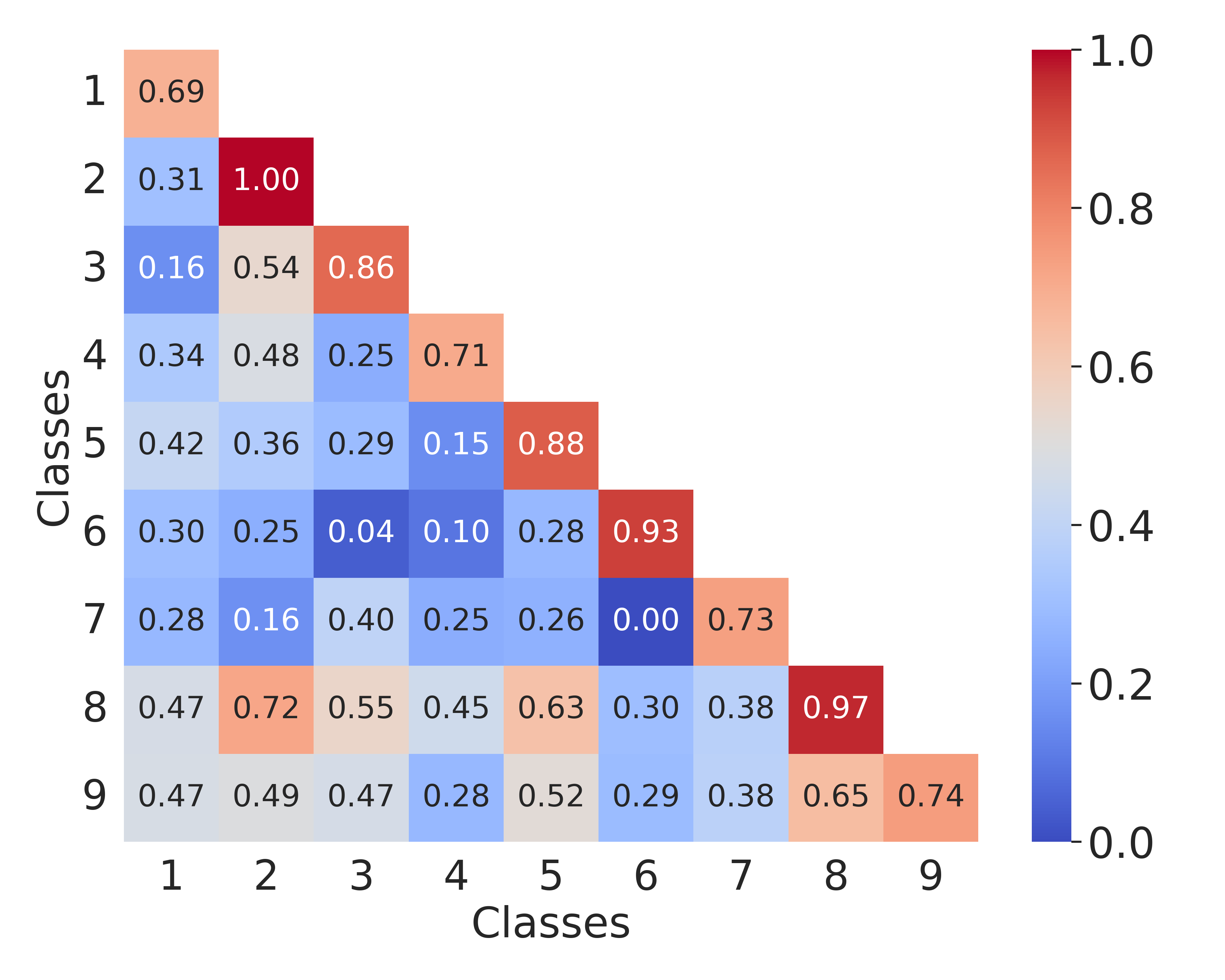}
    \includegraphics[scale = 0.27]{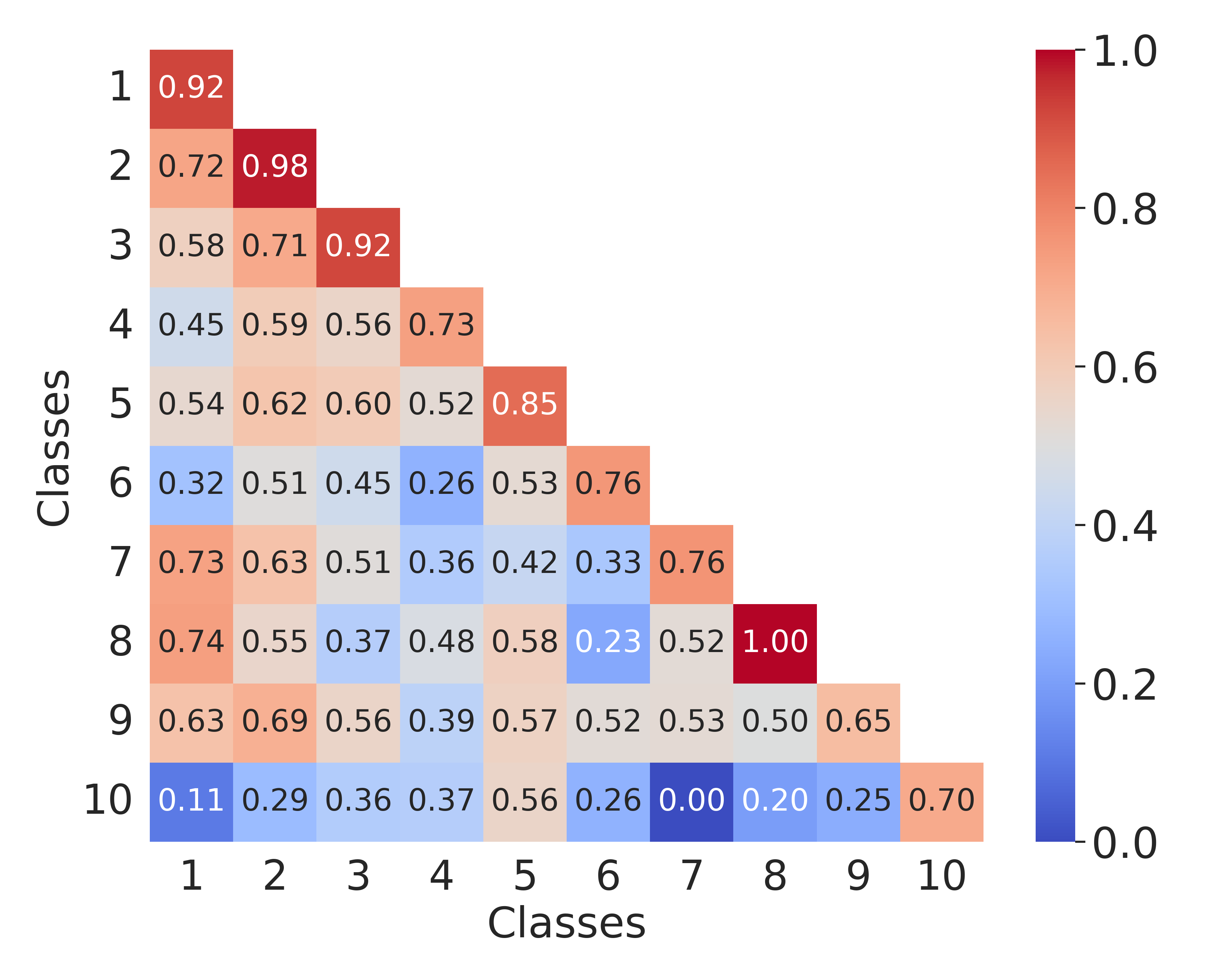}
    \caption{Heatmaps illustrating within-class and between-class cosine similarities of text embeddings derived from the testing time series data in Japanese Vowels \textbf{(left)} with 9 classes and Spoken Arabic Digits \textbf{(right)} with 10 classes. On both axes, x and y represent different classes. Diagonal entries indicate within-class similarities, and off-diagonal entries represent between-class similarities. Warmer colors signify higher cosine similarities, while cooler colors suggest lower similarities.}
    \label{cosine_similarity_val}
\end{figure*}

Figure \ref{cosine_similarity_val} visualizes the within-class and between-class cosine similarities of text embeddings derived from the testing time series. 
Each matrix entry is scaled using min-max normalization to range from 0 to 1, where warmer colors in the heatmap represent higher similarities and darker shades indicate lower similarities. Diagonal entries show within-class similarities, highlighting intra-class cohesion, while off-diagonal entries reveal between-class relationships

Similar to the training set, we observe that within-class similarity consistently exceeds across-class similarity, thereby validating the hypothesis that text embeddings effectively retain and convey significant information from the underlying time series data.

\section{Numerical Precision for Tokenization} \label{appendix:prec_analysis}
\begin{figure*}
    \centering
    \includegraphics[scale = 0.55]{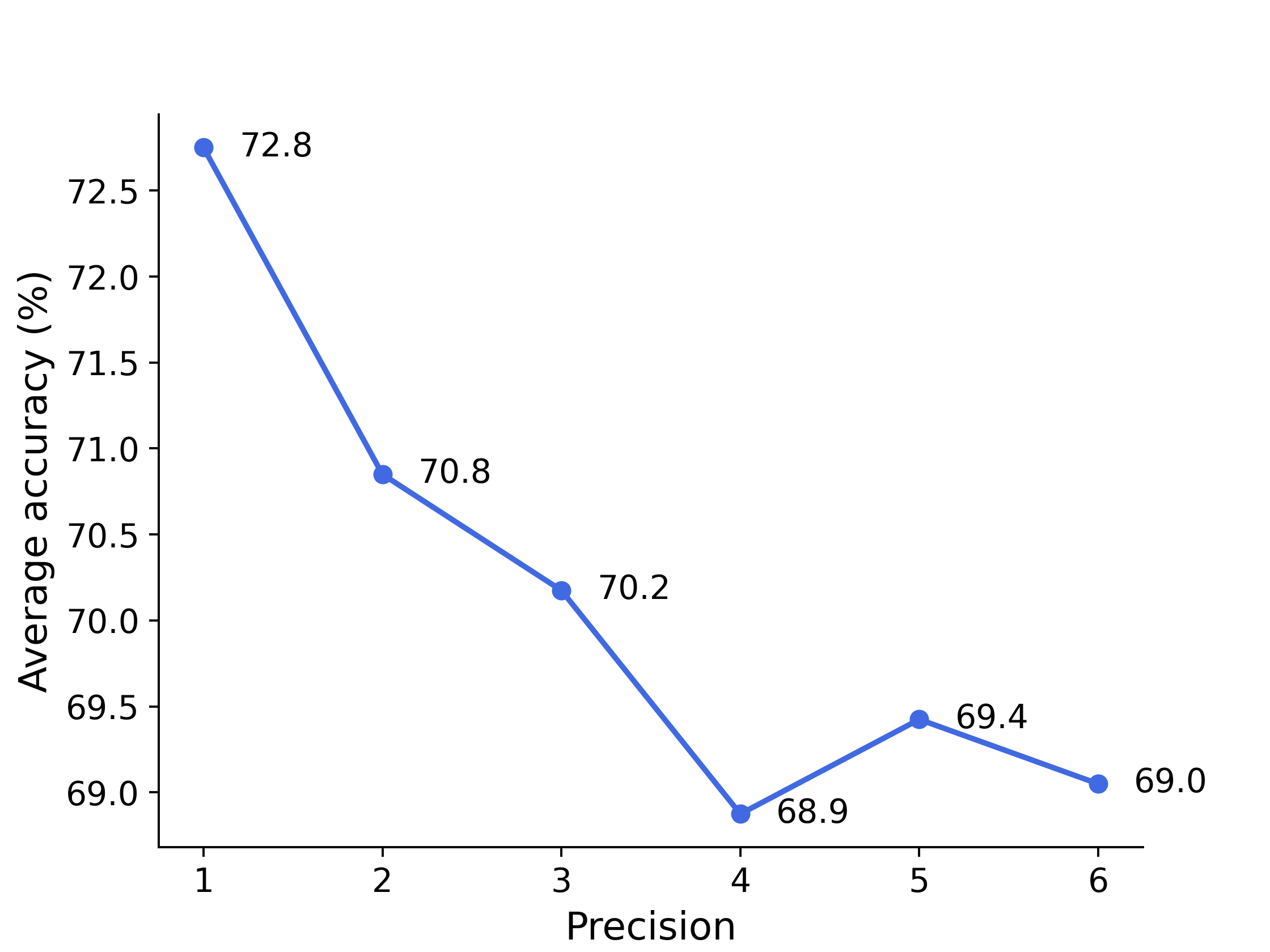}
    \caption{\label{fig:precision_analysis}Average classification accuracy (\%) across numerical precisions 1 to 6. Results are averaged from four datasets: Handwriting, Heartbeat, JapaneseVowels, and UWaveGestureLibrary. See Table \ref{table:appendix_precision_analysis} in the Appendix for detailed results.}
\end{figure*}

To explore the impact of numerical precision on the computation of embeddings, we analyzed classification accuracy across precisions 1 to 6 using four datasets: Handwriting, Heartbeat, JapaneseVowels, and UWaveGestureLibrary. We selected these datasets because they were the smaller ones among the 10 available, making the computation of embeddings more computationally affordable. Figure \ref{fig:precision_analysis} illustrates the average classification accuracy (\%) across these numerical precisions. 
Detailed results can be found in Table \ref{table:appendix_precision_analysis}.
\begin{table*}[htb]
\caption{Numerical Precision for Tokenization: This table reports classification accuracy (\%). \textcolor{red}{\textbf{Red:}} Best performance. }
\label{table:appendix_precision_analysis}
\centering
\begin{tabular}{@{}l|cccccc@{}}
\toprule
Dataset / Precision & 1 & 2 & 3 & 4 & 5 & 6\\ \midrule
Handwriting & 23.2 & 15.9 & 11.6 & 11.1 & 12.0 & 11.2 \\
Heartbeat & 78.0 & 78.5 & 79.0 & 78.5 & 79.5 & 78.0 \\
JapaneseVowels & 99.2 & 98.4 & 97.6 & 96.8 & 95.9 &  95.1 \\
UWaveGestureLibrary & 90.6 & 90.6 & 92.5 & 89.1 & 90.3 & 91.9  \\
\midrule
\cellcolor{pale7} \textbf{Average} $\uparrow$ & \cellcolor{pale7} \textcolor{red}{\textbf{72.75}} & \cellcolor{pale7} 70.85 & \cellcolor{pale7} 70.175 & \cellcolor{pale7} 68.875 & \cellcolor{pale7} 69.425 & \cellcolor{pale7} 69.05 \\
\midrule
\cellcolor{pale7} \textbf{AvgWins \%} $\uparrow$ & \cellcolor{pale7} \textcolor{red}{\textbf{50\%}} & \cellcolor{pale7} 0\% & \cellcolor{pale7} 25\% & \cellcolor{pale7} 0\% & \cellcolor{pale7} 25\% & \cellcolor{pale7} 0\% \\
\bottomrule
\end{tabular}
\end{table*}

Studies in the NLP domain have shown that longer inputs do not perform well with language models \cite{press2020shortformer, levy2024same}. Our empirical analysis supports this claim, revealing a decrease in classification accuracy with increased numerical precision when computing text embeddings. The average accuracy starts at 72.8\% with precision 1 and declines to 69\% at precision 6. Additionally, the percentage of AvgWins is highest at precision 1. As numerical precision increases, so does the length of the time series and the input to the text embedding model. Note that the maximum token length for \texttt{text-embedding-3-large} embeddings is 8191, and thus keeping precision of 1 ensures that context length doesn't exceed the maximum permissible token length. This issue is especially problematic for datasets with longer time series, such as the EthanolConcentration dataset, which includes 1751 time steps per sample. 
This finding led us to opt for precision 1 in our study. 

Note that these precision results also depend on the type of tokenization selected, and defining an appropriate tokenization for time series is one of the potential future directions for this work. 

\section{Generalization Across Various Text Embedding Models}\label{appendix:emb_type_analysis}

To evaluate the generalization capabilities of our approach across different text embedding models beyond \texttt{text-embedding-3-large} \cite{openai_large_embeddings}, the performance of \letsc{} was tested using three alternative embedding models: \texttt{e5-mistral-7b-instruct} \cite{wang-etal-2024-improving-text}, \texttt{gte-large-en-v1.5} \cite{li2023towards}, and \texttt{nomic-embed-text-v1} \cite{nussbaum2024nomic}. A summary of the embedding models utilized in this study is provided in Table \ref{table:embedding_models}.
\begin{table*}[htb]
\centering
\caption{Summary of selected embedding models used in the study.}
\label{table:embedding_models}
\resizebox{\textwidth}{!}{%
\begin{tabular}{c|l|c|c}
\toprule
\textbf{MTEB Rank} & \textbf{Model} & \textbf{Embedding Dimensions} & \textbf{Max Token Length} \\ \midrule
15 & \texttt{text-embedding-3-large} \cite{openai_large_embeddings} & 3072 & 8191 \\ \midrule
6 & \texttt{e5-mistral-7b-instruct} \cite{wang-etal-2024-improving-text} & 4096 & 32768 \\ \midrule
9 & \texttt{gte-large-en-v1.5} \cite{li2023towards} & 1024 & 8192 \\ \midrule
35 & \texttt{nomic-embed-text-v1} \cite{nussbaum2024nomic} & 768 & 8192 \\ \bottomrule
\end{tabular}
}
\end{table*}

\subsection{\texttt{e5-mistral-7b-instruct}}
The \texttt{e5-mistral-7b-instruct} model \cite{wang-etal-2024-improving-text} is based on the pretrained Mistral-7b checkpoint \cite{jiang2023mistral} and was fine-tuned using the RankLLaMA training methodology \cite{ma2024fine}. It employed LoRA \cite{hu2021lora} with a rank of 16 on a mixture of multilingual datasets, which included both synthetic data and a collection of 13 public datasets, yielding approximately 1.8 million examples after sampling. Proprietary LLMs, such as GPT-4, were prompted to generate diverse synthetic data with instructions in multiple languages. Leveraging the strong language understanding capabilities of the Mistral model, \texttt{e5-mistral-7b-instruct} achieved state-of-the-art results across nearly all task categories on the competitive MTEB benchmark. Techniques such as gradient checkpointing, mixed precision training, and DeepSpeed ZeRO-3 were applied to further reduce GPU memory requirements.

\subsection{\texttt{gte-large-en-v1.5}}
The \texttt{gte-large-en-v1.5} model \cite{li2023towards} is a general text embedding (GTE) model that utilizes contrastive learning on an open-source large-scale dataset comprising unsupervised text pairs extracted from various sources. To enhance the quality of the learned text representations, high-quality text pairs with human labels from multiple sources were employed for contrastive fine-tuning. The model utilizes a multi-stage contrastive learning approach and benefits from a diverse training data mixture, enabling it to achieve strong generalization performance for single-vector embeddings.

\subsection{\texttt{nomic-embed-text-v1}}
The \texttt{nomic-embed-text-v1} model \cite{nussbaum2024nomic} is a fully reproducible, open-source English text embedding model with an 8192 context length that outperforms both OpenAI Ada-002 and OpenAI text-embedding-3-small on short and long-context tasks. To accommodate long sequence lengths, the model adapts the BERT architecture \cite{devlin-etal-2019-bert} with several optimizations: replacing absolute positional embeddings with rotary positional embeddings \cite{su2024roformer}, using SwiGLU activation instead of GeLU \cite{shazeer2020glu}, implementing Flash Attention \cite{dao2022flashattention}, setting Dropout to 0 \cite{geiping2023cramming}, and ensuring the vocabulary size is a multiple of 64 \cite{portes2024mosaicbert, shoeybi2019megatron}. These modifications result in a 137M parameter encoder.





\section{Trade-offs: Model Accuracy vs. Parameter Complexity}\label{appendix:tradeoff}
Table \ref{appendixtable:tradeoff_accuracy_parameters} illustrates the trade-off between model accuracy and the complexity of training parameters in our model, which utilizes both embeddings and time series data as inputs to a lightweight framework.
To vary model size, we adjust the number of linear and convolution layers in the classification head, ranging from 1 to 5 layers each, creating model variants of different sizes.

\begin{table*}[t!]
\caption{Trade-off between model accuracy and the complexity of training parameters. The accuracy and parameters of the best model are highlighted in \textbf{bold}. The accuracy difference is calculated as the raw difference between the accuracies of the reduced model and the best model. The \% Delta in accuracy and parameters is defined separately for each as \(100 \times \frac{\text{Accuracy of the reduced model}}{\text{Accuracy of the optimal model}}\) and \(100 \times \frac{\text{Parameters of the reduced model}}{\text{Parameters of the optimal model}}\), quantifying the accuracy and computational efficiency of the reduced model relative to our best model.}
\label{appendixtable:tradeoff_accuracy_parameters}
\centering
\resizebox{0.87\linewidth}{!}{%
\begin{tabular}{@{}l|P{2.4cm}P{2.4cm}|P{3.1cm}P{2cm}@{}}
\toprule
Dataset & Accuracy (\%) $\mathbf{\uparrow}$  & Trainable Parameters $\mathbf{\downarrow}$ & Difference $|$ \% Delta  in Accuracy $\mathbf{\uparrow}$ & \% Delta in Parameters $\mathbf{\downarrow}$ \\
\midrule
\multirow{2}{3.4cm}{EthanolConcentration} 
& \cellcolor{pale7} \textbf{52.9} & \cellcolor{pale7} \textbf{283950} & - & - \\
& 46 & 105344 & -6.9 $|$  86.95 & 37.09 \\
\midrule
\multirow{4}{3.4cm}{FaceDetection}
& \cellcolor{pale7} \textbf{68.9} & \cellcolor{pale7} \textbf{3842} & - & - \\
& 68.6 & 2402 & -0.3 $|$ 99.56 & 62.51 \\
& 67.9 & 962 & -1.0 $|$ 98.54 & 25.03  \\
& 66.4 & 482 & -2.5 $|$ 96.37 & 12.54  \\
\midrule
\multirow{5}{3.4cm}{Handwriting} & \cellcolor{pale7} \textbf{23.8} & \cellcolor{pale7} \textbf{154526} & - & - \\
& 23.2 & 107226 & -0.6 $|$ 97.47 & 69.39 \\
& 22.7 & 53626 & -1.1 $|$ 95.37  & 34.70 \\
& 20.2 & 20394 & -3.6 $|$ 84.87 & 13.19 \\
& 19.4 & 13426 & -4.4 $|$ 81.51 & 8.68 \\
\midrule
\multirow{3}{3.4cm}{Heartbeat}
& \cellcolor{pale7} \textbf{78} & \cellcolor{pale7} \textbf{46426} & - & - \\
& 77.6 & 34820 & -0.4 $|$ 99.48 & 75.00 \\
& 75.1 & 26908 & -2.9 $|$ 96.28  & 57.95 \\
\midrule
\multirow{4}{3.4cm}{Japanese Vowels}
& \cellcolor{pale7} \textbf{99.2} & \cellcolor{pale7} \textbf{148233} & - & - \\
& 98.9 & 123401 & -0.3 $|$ 99.69 & 83.24 \\
& 98.6 & 105353 & -0.6 $|$ 99.39 & 71.07 \\
& 98.4 & 100857 & -0.8 $|$ 99.19 & 68.03 \\
\midrule
\multirow{5}{3.4cm}{PEMS-SF} 
& \cellcolor{pale7} \textbf{93.1}  & \cellcolor{pale7} \textbf{564231} & - & - \\
& 91.3 & 173866 & -1.8 $|$ 98.06 & 30.81 \\
& 90.8  & 85210 & -2.3 $|$ 97.52 & 15.10 \\
& 87.9  & 69077 & -5.2 $|$ 94.41 & 12.24 \\
& 87.3 & 62901 & -5.8 $|$ 93.77 & 11.14 \\
\midrule
\multirow{2}{3.4cm}{Self-Regulation SCP1} 
&  \cellcolor{pale7} \textbf{93.2} &  \cellcolor{pale7} \textbf{302626} & - & - \\
& 92.5 & 99657 & -0.7 $|$ 99.24 & 32.93 \\
\midrule
\multirow{5}{3.4cm}{Self-Regulation SCP2} 
& \cellcolor{pale7} \textbf{62.8}  & \cellcolor{pale7} \textbf{334402} & - & -\\
& 58.9 & 166306 & -3.9 $|$ 93.78  & 49.73 \\
& 57.8 & 111106 & -5.0  $|$ 92.03 & 33.22 \\
& 57.2 & 83330 & -5.6 $|$ 91.08 & 24.91\\
& 56.1 & 76386 & -6.7 $|$ 89.33 & 22.84 \\
\midrule
\multirow{8}{3.4cm}{Spoken Arabic Digits}
& \cellcolor{pale7} \textbf{99.2} & \cellcolor{pale7} \textbf{141790} & - & - \\
& 99 & 70066 & -0.2 $|$ 99.79 & 49.41 \\
& 98.4 & 46658 & -0.8 $|$ 99.19 & 32.90 \\
& 98.1 & 30964 & -1.1 $|$ 98.89 & 21.83 \\
& 98 & 20646 & -1.2 $|$ 98.79 & 14.56 \\
& 97.8 & 15487 & -1.4 $|$ 98.58 & 10.92 \\
& 97.6 & 10328 & -1.6 $|$ 98.38 & 7.28 \\
& 96.6 & 5308 & -2.6 $|$ 97.37 & 3.74 \\
\midrule
\multirow{3}{3.4cm}{UWave Gesture Library} 
& \cellcolor{pale7} \textbf{90.6} & \cellcolor{pale7} \textbf{263338} & - & - \\
& 88.4 & 211556 & -2.2 $|$ 97.57 & 80.33  \\
& 87.5 & 176298 & -3.1 $|$ 96.57 & 66.94 \\
\bottomrule
\end{tabular}
}
\end{table*}

 Across all datasets, we find that significant reductions in model parameters result in only a minimal loss of accuracy. This trade-off between model accuracy and parameter complexity is data-dependent. However, we generally observe a trend where a reduction in parameters leads to only a slight decrease in accuracy. Next, let's take a closer look at three datasets: Heartbeat, PEMS-SF, and Spoken Arabic Digits, to understand how these trade-offs manifest in different contexts.

\begin{itemize}
    \item \textbf{Heartbeat:} In the Heartbeat dataset, we retain 99.48\% of the optimal model’s accuracy using only 75\% of its trainable parameters. Specifically, the optimal model achieves an accuracy of 78\% with 46,426 trainable parameters, while the second-best model achieves 77.6\% accuracy with 34,820 parameters. Moreover, we retain 96.28\% accuracy with a further reduction to 57.95\% of the parameters.

    \item \textbf{PEMS-SF:} In the PEMS-SF dataset, the optimal model starts with an accuracy of 93.1\% and 564,231 trainable parameters. Reducing the parameters to 173,866 (30.81\% of the original), the model maintains 98.06\% of its optimal accuracy at 91.3\%. Further parameter reductions to 85,210 (15.10\%), 69,077 (12.24\%), and 62,901 (11.14\%) result in accuracies of 90.8\%, 87.9\%, and 87.3\%, respectively. These reductions illustrate that even significant reductions in parameters only lead to a slight decrease in performance.

    \item \textbf{SpokenArabicDigits:} For the Spoken Arabic Digits dataset, reducing the number of trainable parameters generally correlates with a minor decline in accuracy, though the trade-off is modest. The optimal model, achieving an accuracy of 99.2\% with 141,790 trainable parameters, shows that even with substantial reductions to 70,066 parameters (49.41\% of the original), the accuracy remains high at 99\%, retaining 99.79\% of the original model's accuracy. Further reductions to 46,658 (32.90\%), 30,964 (21.83\%), 20,646 (14.56\%), 15,487 (10.92\%), 10,328 (7.28\%), and 5,308 (3.74\%) yield accuracies of 98.4\%, 98.1\%, 98\%, 97.8\%, 97.6\%, and 96.6\% respectively.
\end{itemize}
These examples highlight that efficiency in terms of trainable parameters does not linearly correspond to a loss in model accuracy across various datasets. While fewer parameters generally lead to a lower accuracy, the decrement is often proportional and manageable, making these models highly suitable for deployment in resource-constrained environments or for applications requiring rapid processing with minimal computational overhead.

\section{Ablation Study}\label{appendix:ablation_study}
Table \ref{table:ablation_embeddings} presents the results of this study, comparing the performance of our default approach, which adds embeddings to time series, to variants that exclude either embeddings or the time series.
\begin{table*}[htb]
\caption{Comparison of classification accuracy (\%) for different configurations: ours (both embeddings and time series), embeddings only, and time series only. \textcolor{red}{\textbf{Red}}: Best.}
\label{table:ablation_embeddings}
\centering
\resizebox{\linewidth}{!}{%
\begin{tabular}{@{}ll|cccccccccc|c|c@{}}
\toprule
& Method/Dataset & EC & FD & 
HW & 
HB & JV & PEMS-SF & SCP1 & SCP2 & SAD & UW 
& 
\cellcolor{pale7} \textbf{Average} $\mathbf{\uparrow}$  & 
\cellcolor{pale7} \textbf{AvgWins \%} $\mathbf{\uparrow}$ 
\\ 
\midrule
& \letsc{} (Ours) & 52.9 & 68.9
& 23.8
& 78   & 99.2 & 93.1 & 93.2 & 62.8 & 99.2 & 90.6 & 
\cellcolor{pale7} \textcolor{red}{\textbf{76.17}} & \cellcolor{pale7} \textcolor{red}{\textbf{60\%}}
\\
\midrule
& \quad embedding only & 38 & 59.4 
& 20.1
& 80  & 99.2 & 92.5 & 88.7 & 63.9 & 99.2 & 
88.4 & 
\cellcolor{pale7} 72.94 & \cellcolor{pale7} 40\%
\\
& \quad time series only & 42.6 & 69.6 & 25.1 & 76.1  & 98.1 & 89 & 93.2 & 58.3 & 98.9 & 83.8 & 
\cellcolor{pale7} 73.47 & \cellcolor{pale7} 30\%
\\ 
\bottomrule
\end{tabular}
}
\end{table*}

\section{Alternative Methods for Fusing Time Series with Embeddings}\label{appendix:alternative_fusion_methods}
We explored two additional methods for fusing time series and embeddings beyond simple addition. The first method involves a \textit{Fusion network} that first processes embeddings and time series data through convolutional and dense layers in two separate branches, then merges the features from both branches into a final dense network. The second method employs \textit{Concatenation}, where the time series and embeddings are concatenated and processed through a lightweight classification head. Despite cross-attention being another alternative for fusing different modalities, we didn't include it in this study due to the computational complexity it adds to the model.
Table \ref{table:alternative_methods_integrating} presents the classification accuracy and trainable model parameters for these variations.
\begin{table*}[htb]
\caption{Comparison of classification accuracy (\%) and trainable model parameters (millions) for alternative methods of fusing time series with embeddings. Higher AvgWins and accuracy averages indicate superior performance, while lower model parameter averages suggest greater computational efficiency. \textcolor{red}{\textbf{Red:}} Best performance.
}
\label{table:alternative_methods_integrating}
\centering
\resizebox{\linewidth}{!}{%
\begin{tabular}{@{}l|cccccccccc|c|c@{}}
\toprule
\multicolumn{13}{c}{\bf Accuracy	$\mathbf{\uparrow}$}\\
\midrule
Method/Dataset & EC & FD & 
HW & 
HB & JV & PEMS-SF & SCP1 & SCP2 & SAD & UW 
& 
\cellcolor{pale7} \textbf{Average}
$\mathbf{\uparrow}$
& 
\cellcolor{pale7} \textbf{AvgWins \%} 
$\mathbf{\uparrow}$ 
\\ 
\midrule
\letsc{} (Addition) & 52.9 & 68.9
& 23.8
& 78   & 99.2 & 93.1 & 93.2 & 62.8 & 99.2 & 90.6 & 
\cellcolor{pale7} \textcolor{red}{\textbf{76.17}} & \cellcolor{pale7} \textcolor{red}{\textbf{70\%}}
\\
\midrule
Fusion network & 44.1 & 66.5
& 23.6
& 76.6   & 98.1 & 86.1 & 92.8 & 56.1 & 
99.2 & 90.9 &
\cellcolor{pale7} 73.40 & \cellcolor{pale7} 20\%
\\
Concatenation & 43 & 65.1 & 22.5 & 79  & 98.9 & 93.1 & 93.9 & 58.9 & 99 & 88.8 & 
\cellcolor{pale7} 74.22 & \cellcolor{pale7} 30\%
\\ 
\bottomrule
\toprule
\multicolumn{13}{c}{\bf Trainable Parameters (M) $\mathbf{\downarrow}$} \\
\midrule
Method/Dataset & EC & FD & 
HW & 
HB & JV & PEMS-SF & SCP1 & SCP2 & SAD & UW 
& 
\multicolumn{2}{c}{ \cellcolor{pale7} \textbf{Average} $\mathbf{\downarrow}$ }\\
\midrule
\letsc{} (Addition) & 0.28 & 0.003 & 0.15 & 0.04 & 0.14 & 0.56 & 0.30 & 0.33 & 0.14 & 0.26 & \multicolumn{2}{c}{ \cellcolor{pale7} \textcolor{red}{\textbf{0.22}} }\\
\midrule
Fusion Network & 0.19 &  0.35 & 0.33 & 0.28 & 0.16 & 5.54 & 0.37 & 0.27 & 0.23 & 0.04 & \multicolumn{2}{c}{ \cellcolor{pale7} 0.78 } \\
Concatenation & 0.42 & 0.009 & 0.26 & 0.17 & 0.11 & 0.28 & 0.23 & 0.39 & 0.09 & 0.46 & \multicolumn{2}{c}{ \cellcolor{pale7} 0.24 }\\
\bottomrule
\end{tabular}
}
\end{table*}

We observe that the addition approach in the LETS-C architecture achieves the highest average classification accuracy (76.11\%) compared to the fusion network (73.40\%) and concatenation (74.22\%). Notably, when compared to all baselines, concatenation emerges as the second-best method after addition, which achieves the highest accuracy. The fusion approach ranks fifth overall, with only OneFitsAll and TimesNet outperforming it, following addition and concatenation, in terms of average accuracy.

\end{document}